\documentclass[runningheads]{llncs}

\usepackage{eccv}

\usepackage{eccvabbrv}

\usepackage{graphicx}
\usepackage{booktabs}
\usepackage{marvosym}
\usepackage{lipsum}
\newcommand\blfootnote[1]{%
	\begingroup
	\renewcommand\thefootnote{}\footnote{#1}%
	\addtocounter{footnote}{-1}%
	\endgroup
}
\usepackage{multirow}
\usepackage{amsmath}
\usepackage{amssymb}
\usepackage{xcolor}  
\usepackage{colortbl} 
\usepackage[accsupp]{axessibility}  

\usepackage{hyperref}

\usepackage{orcidlink}

\begin{document}

\title{MoAKE: Toward Unified All-in-One Action Quality Assessment via Mixture of Action Knowledge Experts} 

\titlerunning{MoAKE: Toward Unified All-in-One Action Quality Assessment}

\author{Huangbiao Xu\inst{1,2}\orcidlink{0000-0002-3717-8713} \and
Huanqi Wu\inst{1,2}\orcidlink{0009-0008-4518-3273} \and
Xiao Ke\inst{1,2,}\textsuperscript{\Letter}\orcidlink{0000-0001-9059-5391} \and
Jiaxin Cai\inst{1}\orcidlink{0009-0003-5697-1746} \and \\
Junyi Wu\inst{1,2}\orcidlink{0000-0002-2509-1223} \and
Jinglin Xu\inst{3}\orcidlink{0000-0002-1553-5441}}

\authorrunning{H. Xu et al.}

\institute{Fujian Provincial Key Laboratory of Networking Computing and Intelligent Information Processing, College of Computer and Data Science, Fuzhou University, Fuzhou 350108, China \and
Engineering Research Center of Big Data Intelligence, Ministry of Education, Fuzhou 350108, China \and
School of Intelligence Science and Technology, University of Science and Technology Beijing, Beijing 100083, China\\
\email{kex@fzu.edu.cn, \{huangbiaoxu.chn, wuhuanqi135, xujinglinlove\}@gmail.com, jiaxincai528@163.com, junyi.wu-1@outlook.com}
\blfootnote{\textsuperscript{\Letter} Corresponding Author.}}
\maketitle

\begin{abstract}
  Action Quality Assessment (AQA) aims to objectively evaluate performance quality from action videos. Most existing methods follow a ``one-by-one'' paradigm, training a separate model for each action type. This setting limits real-world deployment, as it requires prior action-type knowledge to select the corresponding model and suffers from poor generalization across diverse actions. To address these limitations, we study the challenging task of \textbf{all-in-one AQA}, which aims to assess heterogeneous actions within a single unified model. We propose a novel \textbf{Mixture of Action Knowledge Experts (MoAKE)} framework, designed to mitigate negative knowledge transfer caused by large semantic discrepancies among actions. MoAKE learns complementary experts that capture diverse action patterns within a shared semantic space and dynamically aggregates their knowledge to adapt the assessment to the input action. Each expert is tailored with segment-aware prototypes to handle varying temporal lengths, together with an \textbf{Adaptive Intra- and Inter-Segment Relationship Modeling (AIISRM)} module to model multi-granularity temporal dynamics. Furthermore, we establish \textbf{comprehensive benchmarks} for \textbf{all-in-one} as well as \textbf{zero/few-shot AQA}. Extensive experiments on three long-term datasets demonstrate that MoAKE significantly outperforms existing methods in the all-in-one setting, while also achieving consistent generalization on three short-term datasets under zero/few-shot evaluation. Code is available at https://github.com/XuHuangbiao/MoAKE.
\end{abstract}

\section{Introduction}
Action Quality Assessment (AQA) has made significant strides in recent years across various domains such as sports \cite{xu2024fineparser,11024123,xu2024procedure,pami/XuYP25}, medical rehabilitation \cite{liu2021towards,10049714,bruce2024egcn++}, and skill determination \cite{Yu_2026_CVPR,xu2025dancefix}. The primary goal of AQA is to learn a mapping from motion semantics to expert-provided scores. To tackle this challenging problem, recent research often attempts to model fine-grained semantics \cite{zhou2023hierarchical,xu2024fineparser,xu2024procedure,pami/XuYP25} or incorporate additional semantic cues from other modalities \cite{xu2024vision,Xu_2025_CVPR,tip/ZengZ24,xia2023skating,xu2025quality}. However, these advances are mostly developed under per-action settings.

\begin{figure}[!t]
	\centering
	\includegraphics[width=\linewidth]{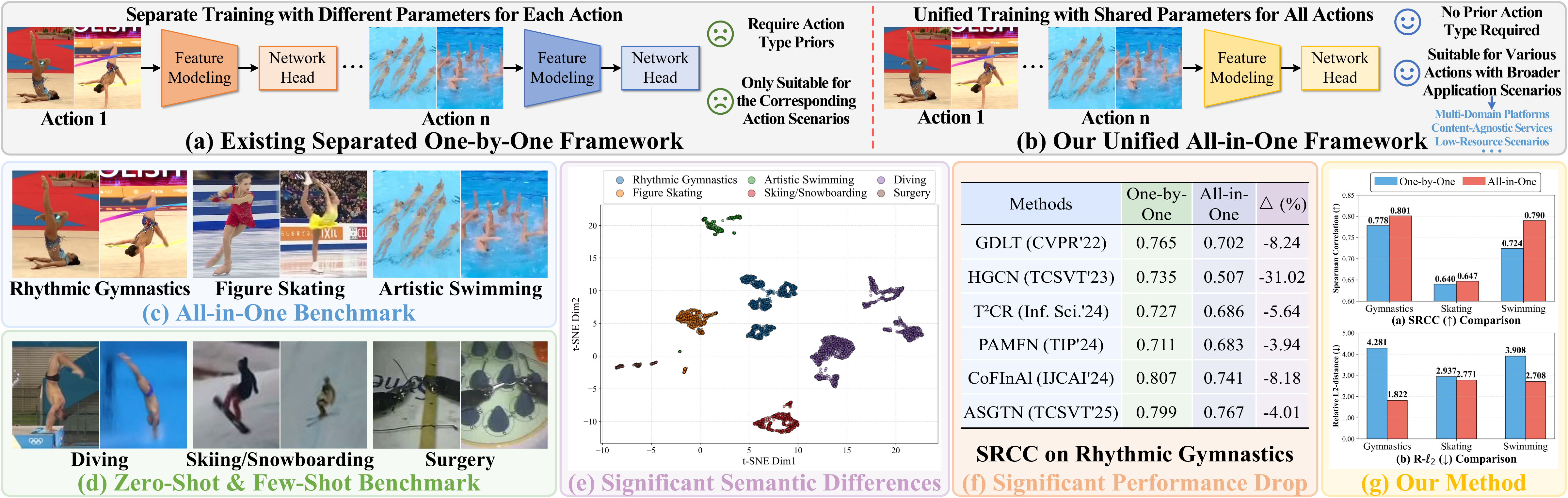}
    \caption{Illustration of paradigm comparisons, benchmarks, and challenges. (a) Existing separated one-by-one framework trains specific models for each action type. (b) Our unified all-in-one framework assesses heterogeneous actions using a single model. (c) Our all-in-one AQA benchmark. (d) Our zero/few-shot AQA benchmark. (e) Significant semantic differences across actions pose a major challenge. (f) Naively training all-in-one models leads to negative transfer. (g) Our method enables positive transfer.}
	\label{fig:1}
\end{figure}
Despite their success, the prevailing approaches remain strictly confined to a ``one-by-one'' paradigm (Fig.~\ref{fig:1} (a)), where separate models are trained and deployed for each specific action type. While effective in isolated benchmarks, this fragmentation creates major barriers to real-world scalability across various applications. \emph{For example}: \emph{first}, it relies on reliable metadata (\ie, known action types) to select the correct model, complicating deployment in \textbf{open-domain platforms}~\cite{hu2023exploring} where diverse user-uploaded content lacks reliable labels; \emph{second}, it hinders \textbf{content-agnostic services}~\cite{kayal2025large} that must process heterogeneous or even mixed actions without manual intervention; and \emph{third}, it suffers in \textbf{low-resource scenarios}~\cite{nc2022addressing}, where training and maintaining high-capacity models for new or rare actions is data- and compute-intensive. Motivated by these practical needs, and inspired by prior AQA studies~\cite{parmar2019action,xu2024procedure,xu2025quality} that suggest the value of cross-action transfer, we study a more generalizable paradigm: \textbf{All-in-One AQA} (Fig.~\ref{fig:1} (b)), where a single unified model assesses heterogeneous actions.

However, the path to an all-in-one AQA model is fraught with difficulties. The core challenge stems from the substantial semantic discrepancies across various actions, as illustrated in Fig. \ref{fig:1} (e). Directly applying existing methods to all-in-one training often leads to ``\textbf{negative transfer},'' where knowledge from disparate actions interferes with each other~\cite{jiang2023forkmerge}. This results in significant performance degradation (an average SRCC drop of 10.17\%), as shown in Fig. \ref{fig:1} (f). This decline suggests that existing models lack mechanisms to adaptively identify relevant assessment patterns for each input action and fail to leverage the rich, albeit varied, information from multiple domains. However, these diverse action cues can be complementary when properly utilized. As shown in Fig. \ref{fig:1} (g), our all-in-one approach effectively mixes multi-action knowledge and demonstrates positive transfer (details in \cref{sec:SOTA}). \emph{The key insight, therefore, is to transform this challenge into an opportunity by designing an all-in-one model that can extract and synthesize complementary knowledge from various actions.}

To this end, we propose the \textbf{M}ixture \textbf{o}f \textbf{A}ction \textbf{K}nowledge \textbf{E}xperts (\textbf{MoAKE}), a novel framework designed to mitigate negative transfer and enable adaptive assessment. Instead of forcing a single, monolithic representation, MoAKE learns a set of discriminative action knowledge experts, each specializing in capturing complementary action patterns and mapping them to a shared semantic space. Within each expert, fixed-length, segment-aware prototypes are utilized to learn and aggregate segment semantics, adapting to action scenarios of varying lengths. This is complemented by a novel \textbf{A}daptive \textbf{I}ntra- and \textbf{I}nter-\textbf{S}egment \textbf{R}elationship \textbf{M}odeling (AIISRM) module. AIISRM empowers the experts to capture multi-granularity relationships and adaptively model segment-level patterns, transforming common knowledge into action-specific characteristics. By dynamically mixing the insights from different experts, MoAKE tailors its evaluation to the specific action instance, effectively mitigating knowledge interference and enabling robust score predictions.

To demonstrate the effectiveness and generalization capabilities of our approach, we establish new comprehensive benchmarks, including both all-in-one and zero/few-shot AQA tasks. As depicted in Fig. \ref{fig:1} (c)-(d), we train a unified model on three mainstream long-term action types (rhythmic gymnastics~\cite{zeng2020hybrid}, figure skating~\cite{16}, and artistic swimming~\cite{zhang2023logo}) and evaluate its zero/few-shot generalization capabilities on three short-term actions (diving~\cite{parmar2019and}, snowboarding/skiing~\cite{parmar2019action}, and surgery~\cite{jigsaws}). Extensive experiments and ablation studies reveal the importance of leveraging the complementarity of different action knowledge for all-in-one AQA, demonstrating that our method outperforms state-of-the-art methods. This work, to our knowledge, pioneers a new direction by enabling the assessment of multiple action types within an all-in-one framework.

Our main contributions are summarized as follows:
\begin{itemize}
  \renewcommand{\labelitemi}{$\bullet$}
	\item We propose a novel Mixture of Action Knowledge Experts framework for all-in-one AQA, which dynamically fuses knowledge from multiple experts to bridge the semantic gap between diverse action types, adapting its assessment to the input action.
	\item We introduce action knowledge experts that model representative action patterns within a shared semantic space, and an Adaptive Intra- and Inter-Segment Relationship Modeling module to dynamically capture crucial segment-aware contextual correlations.
	\item We establish a novel comprehensive benchmark for all-in-one and zero/few-shot AQA, validating the state-of-the-art performance of our method on six mainstream long-term and short-term action datasets in both unified assessment and cross-action generalization.
\end{itemize}

\section{Related Work}
\subsection{Action Quality Assessment}
Action Quality Assessment (AQA) aims to learn a complex mapping from motion semantics to expert evaluation. To address this, recent research often follows two directions: fine-grained semantic modeling, which enhances discriminative power by focusing on finer sub-actions \cite{xu2022finediving,xu2024fineparser,pami/XuYP25,bai2022action}, hierarchical structures \cite{zhou2023hierarchical,ke2024two}, uncerainty modeling~\cite{17}, or discrete quality grades \cite{xu2022likert,10884538,CoFInAl,11024123}; and multi-modal learning, which leverages auxiliary information from audio \cite{xia2023skating}, flow \cite{tip/ZengZ24,zhou2026brima}, text \cite{Xu_2025_CVPR,xu2025quality,majeedi2024rica}, or pose \cite{bruce2024egcn++}. As multimodal learning~\cite{cai2025keep,cai2024efficient,wu2025enhanced,HuangSXK25,11164481,wang2026multimodal,feng2026training,zhang2026multimodal} has become increasingly mature, some studies~\cite{xu2026mcmoe,xu2026limssr} have begun to explore incomplete multimodal AQA that better reflects real-world scenarios. Specifically, MCMoE~\cite{xu2026mcmoe} pioneers the adoption of a mixture-of-experts (MoE)~\cite{li2026integrating,li2025ipcmoe} approach to investigate stable AQA models when modalities are missing. Meanwhile, LIMSSR~\cite{xu2026limssr} further explores the challenging scenario of training without access to complete modal data. 

However, these methods adhere to a ``one-by-one'' paradigm, training specialized models for individual action types. This limits real-world applicability, as it requires prior knowledge of the action type and struggles to generalize across significant semantic gaps. Consequently, existing methods are difficult to deploy in scenarios with unknown action types, such as open-domain platforms, content-agnostic services, and low-resource environments. \emph{In contrast, our MoAKE is designed for challenging all-in-one AQA, dynamically mixing knowledge from multiple specialized experts to adaptively assess diverse actions within a unified model.}

\subsection{All-in-One Models}
The ``all-in-one'' model, which handles multiple tasks or domains with a single set of weights, has gained widespread attention for its efficiency and generality in fields like image restoration \cite{potlapalli2023promptir,ai2024multimodal,li2022all,11123156}, motion understanding \cite{zhou2024avatargpt,li2024all}, and vision-language learning \cite{10356651,wang2023all}. However, this paradigm presents unique challenges for AQA. Unlike other tasks, AQA requires sensitivity to subtle execution details that significantly impact scores. This difficulty is amplified when a single model must contend with the vast semantic differences across action types. Thus, existing all-in-one methods are ill-suited for AQA because they are not designed to navigate the nuanced quality assessment space. \emph{To our knowledge, our MoAKE represents an early attempt to systematically explore all-in-one and zero/few-shot AQA, leveraging a mixture of discriminative experts to adapt to different assessment patterns and effectively utilize cross-domain knowledge.}

\begin{figure}[!t]
	\centering
	\includegraphics[width=\linewidth]{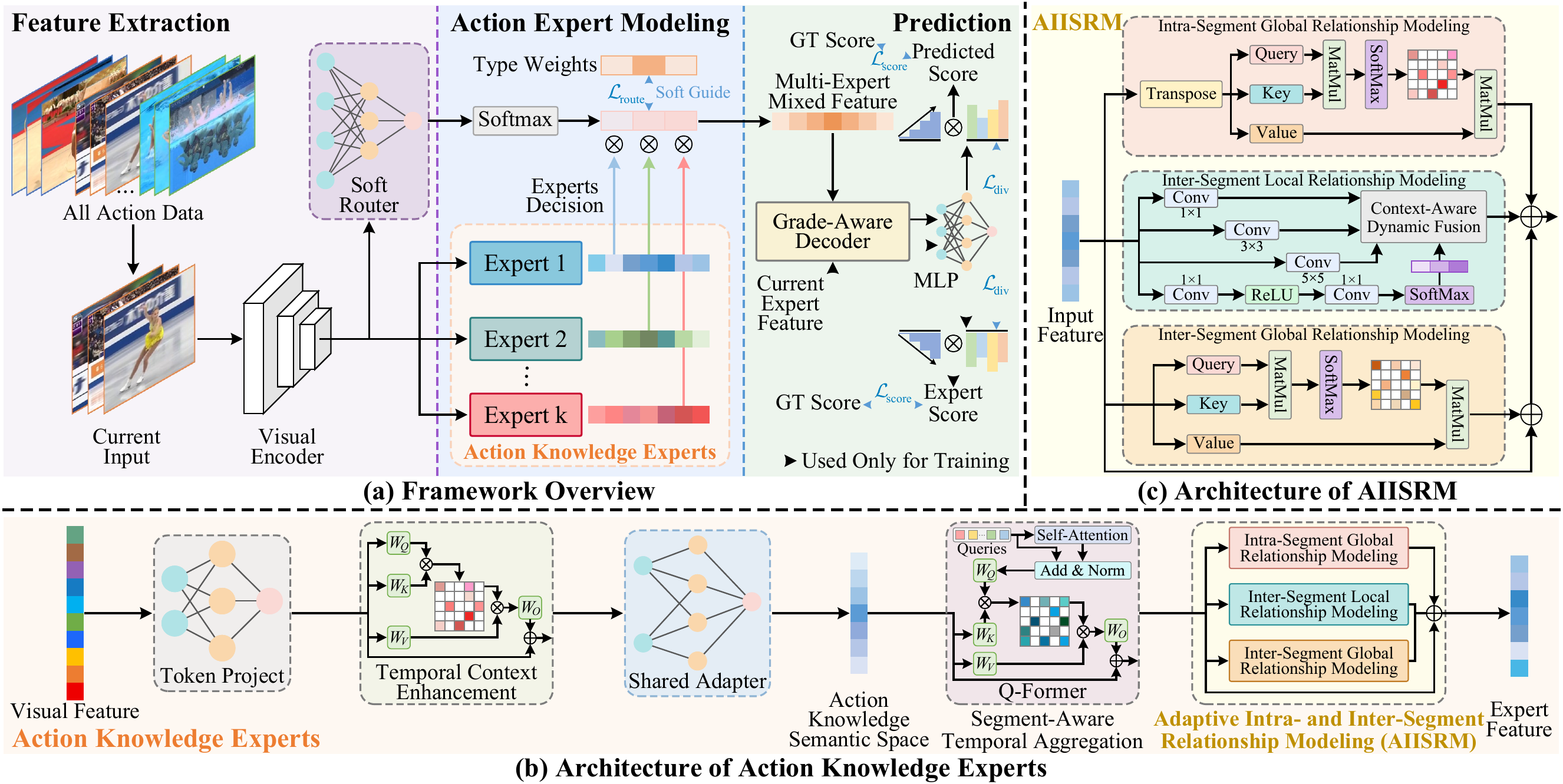}
     \caption{Overview of our MoAKE framework. (a) MoAKE processes video features via multiple action knowledge experts, dynamically fusing their outputs for score prediction. (b) Each expert maps features to a shared space, aggregates semantics using segment-aware prototypes, and refines them via our AIISRM. (c) AIISRM adaptively models intra- and inter-segment relationships to capture multi-granularity context.}
	\label{fig:2}
\end{figure}
\section{Method}
In this section, we detail our Mixture of Action Knowledge Experts (MoAKE) framework, designed for the challenging all-in-one AQA, as shown in~\cref{fig:2}.

\subsection{Overview}
The primary challenge in all-in-one AQA is assessing diverse actions with significant semantic differences using a single model. Naively training on diverse actions often causes negative transfer, where conflicting action patterns hinder performance. To address this, our MoAKE framework (\cref{fig:2} (a)) leverages a Mixture-of-Experts (MoE) architecture. \emph{Unlike traditional MoE designs used primarily for parameter scaling via token-level sparse routing~\cite{shazeer2017outrageously,fedus2022switch,riquelme2021scaling}, our MoAKE is specifically tailored for AQA.} First, it operates at the macro action level with a lightweight design to maintain practical computational costs for real-world deployment (as evidenced in \cref{tab:5}). Second, instead of hard routing that strictly isolates knowledge, we employ a soft routing mechanism. This partitions the complex problem space into manageable sub-problems while adaptively blending complementary knowledge, effectively transforming the semantic gap into positive transfer.

Our MoAKE is trained on a consolidated dataset comprising $K$ different action types. For a given input video, we extract a sequence of visual features $\mathbf{X} \in \mathbb{R}^{T \times D}$ using a pre-trained visual encoder, where $T$ is the number of snippets and $D$ is the feature dimension. Note that $T$ varies across different action types due to their temporal characteristics. These features are then processed in parallel by $K$ distinct action knowledge experts $\{ {E_k}\}_{k = 1}^K$, each designed to capture unique discriminative semantic patterns.

Concurrently, a lightweight soft router, $R(\cdot)$, dynamically computes a set of weights based on the features $\mathbf{X}$, determining the contribution of each expert. Finally, action-adaptive multi-expert mixed feature $\mathbf{F}_{\text{mix}}$ is obtained by a weighted sum of the individual expert outputs $\mathbf{F}_k\!=\!E_k(\mathbf{X})$:
\begin{equation}
    \small
    \{ {w_k}\}_{k = 1}^K = R(\mathbf{X}), \quad \mathbf{F}_{\text{mix}} = \sum_{k=1}^{K} w_k \mathbf{F}_k,
    \label{eq3}
\end{equation}
where ${w_k}$ is the expert weights produced by the soft router, satisfying $\sum_k w_k = 1$. This mechanism allows the model to adaptively emphasize the most relevant expert knowledge for the specific action being evaluated.

For score prediction, we adopt a grade-aware decoder inspired by recent grade-based regressors~\cite{xu2022likert,CoFInAl,Xu_2025_CVPR,10884538,11024123}. The decoder, $De(\cdot)$, uses a set of learnable grade queries to transform the mixed feature $\mathbf{F}_{\text{mix}}$ into $M$ grade-aware representations $\{ {\boldsymbol{r}_m}\}_{m = 1}^M$, which is then mapped to a final score $\hat{y}$.
\begin{equation}
    \small
    \{ \boldsymbol{r}_m \}_{m=1}^M = De(\mathbf{F}_{\text{mix}}), \quad \hat{y} = \text{MLP}(\{ \boldsymbol{r}_m \}_{m=1}^M).
    \label{eq:decoder}
\end{equation}

To ensure that each expert learns discriminative knowledge, we encourage each expert to learn score predictions for corresponding actions during training. The entire framework is optimized end-to-end with a composite loss function:
\begin{equation}
    \small
    \mathcal{L}_{\text{total}} = \lambda_1 \mathcal{L}_{\text{score}} + \lambda_2 \mathcal{L}_{\text{route}} + \lambda_3 \mathcal{L}_{\text{div}},
    \label{eq:loss}
\end{equation}
where $\mathcal{L}_{\text{score}}$ is the primary score regression loss, $\mathcal{L}_{\text{route}}$ is a soft routing loss to guide the router using ground-truth action category labels, and $\mathcal{L}_{\text{div}}$ is a diversity loss to ensure different grade patterns focus on distinct action quality. The hyperparameters $\lambda_1$, $\lambda_2$ and $\lambda_3$ balance these objectives.

\subsection{Action Knowledge Experts}
The core of our MoAKE lies in the action knowledge experts. Crucially, unlike standard MoE frameworks where experts are typically simple feed-forward networks~\cite{shazeer2017outrageously,fedus2022switch,riquelme2021scaling}, our experts are custom-designed to handle the unique temporal demands of AQA. As illustrated in \cref{fig:2} (b), each expert $E_k$ is responsible for learning a distinct subset of action patterns. A key challenge is to create a unified representation space that can handle the diverse temporal lengths ($T_1, T_2, \dots, T_K$) and semantic differences of actions. To this end, each expert employs a tailored two-stage process: (1) adaptive temporal aggregation using segment-aware prototypes to normalize variable-length inputs, and (2) multi-granularity relationship modeling via our specifically proposed AIISRM module.

Initially, the input features $\mathbf{X} \in \mathbb{R}^{T \times D}$ are projected into a lower-dimensional space $\mathbb{R}^{T \times d}$ and enhanced with a self-attention layer \cite{vaswani2017attention} to capture initial temporal context. To bridge the gap between different action types, the features are then mapped into a shared semantic space (denoted as $\mathbf{X}' \in \mathbb{R}^{T \times d}$) using a shared adapter, which consists of a two-layer multilayer perceptron (MLP) whose weights are shared across all experts. This encourages the experts to learn a common language for action quality.

\noindent \textbf{Segment-Aware Temporal Aggregation.} To handle variable-length action videos, we introduce a segment-aware temporal aggregation (SATA) mechanism employing $N$ learnable, segment-aware prototypes $\mathbf{P}_k\!\in\!\mathbb{R}^{N \times d}$ specific to each expert $E_k$. Unlike shared ones, each expert utilizes specialized prototypes to capture action-specific temporal patterns and semantic characteristics. This design allows each expert to develop domain-specific knowledge for different action types while processing variable-length inputs. These prototypes $\mathbf{P}_k$ act as learnable queries to distill salient information from $\mathbf{X}'$ into fixed-length representations. Inspired by Q-Former~\cite{li2023blip}, this process uses cross-attention where the ``queries'' are $\mathcal{Q} = \mathbf{W}_\mathcal{Q}^\text{seg} \mathbf{P}_k$ and the ``keys/values'' are $\mathcal{K} = \mathbf{W}_\mathcal{K}^\text{seg} \mathbf{X}', \mathcal{V} = \mathbf{W}_\mathcal{V}^\text{seg} \mathbf{X}'$:\begin{equation}
    \small
    \mathbf{F}_\text{seg}^k = \text{Softmax}\left( \mathcal{Q} (\mathcal{K})^{\mathrm{T}}/{\sqrt{d}} \right) \mathcal{V} \in \mathbb{R}^{N \times d},
    \label{eq:qformer}
\end{equation}
where $\sqrt d$ is a normalization factor, and $\mathbf{W}_\mathcal{Q}^\text{seg}$, $\mathbf{W}_\mathcal{K}^\text{seg}$, $\mathbf{W}_\mathcal{V}^\text{seg}$ are learnable weights. $\mathbf{F}_\text{seg}^k$ is a fixed-length sequence of $N$ tokens summarizing the entire action regardless of its original duration $T$. This standardized representation makes subsequent modules robust to temporal variations.

\noindent \textbf{Expert-Specific Feature Refinement.} The above temporal aggregation produces discriminative fixed-length representations $\mathbf{F}_\text{seg}^k$ that capture action-specific patterns. However, effective AQA requires understanding complex temporal relationships at multiple granularities - both within individual action segments (intra-segment) and across different segments (inter-segment). These relationships are crucial for assessing short-term execution details and long-term action coherence \cite{CoFInAl,zhou2023hierarchical}. To address this need, we introduce the Adaptive Intra- and Inter-Segment Relationship Modeling (AIISRM) module for each expert $E_k$.

AIISRM processes the $\mathbf{F}_\text{seg}^k$ to capture multi-granularity contextual correlations while incorporating residual connections to preserve discriminative characteristics captured by the segment-aware prototypes and stabilize training. The final expert output $\mathbf{F}_k$ is computed as:
\begin{equation}
    \small
    \mathbf{F}_k = \text{AIISRM}_k(\mathbf{F}_\text{seg}^k).
    \label{eq:expert_output}
\end{equation}

\subsection{Adaptive Segment Relationship Modeling}
As introduced, the AIISRM module refines the fixed-length segment-aware features $\mathbf{F}_\text{seg}^k$ by capturing complex temporal dependencies from multiple perspectives. To achieve this, AIISRM, as depicted in \cref{fig:2} (c), innovatively processes its input through three parallel branches, each targeting a distinct relationship. The effectiveness of this multi-branch design is validated in our ablation studies.

\noindent \textbf{Intra-Segment Global Relationship Modeling.} To capture the temporal evolution patterns from a feature-centric perspective, this branch models global correlations along the feature dimension. We first transpose the input features to $(\mathbf{F}_\text{seg}^k)^{\mathrm{T}} \in \mathbb{R}^{d \times N}$ and then apply a standard self-attention mechanism. This allows the model to learn how the values of individual semantic attributes (e.g., velocity, posture cues) evolve and interrelate across the entire sequence of segments. By focusing on these intra-segment temporal dynamics, this branch captures short-term execution details and a complementary view of long-term action patterns:
\begin{equation}
    \small
    \mathbf{F}_\text{intra}^k\!=\!\text{Softmax}\!\left(\!\frac{\mathbf{W}_\mathcal{Q}^\text{intra}(\mathbf{F}_\text{seg}^k)^{\mathrm{T}} \mathbf{W}_\mathcal{K}^\text{intra}\mathbf{F}_\text{seg}^k}{\sqrt{d}}\!\right)\!\mathbf{W}_\mathcal{V}^\text{intra}(\mathbf{F}_\text{seg}^k)^{\mathrm{T}},
\end{equation}
where $\mathbf{W}_\mathcal{Q}^\text{intra}$, $\mathbf{W}_\mathcal{K}^\text{intra}$, and $\mathbf{W}_\mathcal{V}^\text{intra}$ are learnable matrices.

\noindent \textbf{Inter-Segment Local Relationship Modeling.} Assessed actions in AQA often consist of multiple sub-actions, making the local temporal correlations between them complex. To address this, this branch employs convolutions with varying kernel sizes to dynamically model the inter-segment local relationships. Specifically, we use three parallel 1D convolutions with kernel sizes of 1, 3, and 5 to extract multi-scale local features. A lightweight attention mechanism, implemented with a two-layer convolutional block, generates context-aware weights to dynamically fuse these features. This allows the model to adaptively focus on the most relevant local patterns for the assessed actions. Formally,
\begin{gather}
  \small
  \mathbf{F}_{\text{local}, i}^k = \text{Conv1D}_i(\mathbf{F}_\text{seg}^k), i \in \{1, 3, 5\}, \\
  \small
  \boldsymbol{w}_i = \text{Softmax}\left(\text{ConvBlock}(\mathbf{F}_\text{seg}^k)\right), \\
  \small
  \mathbf{F}_\text{local}^k = \sum_{i \in \{1, 3, 5\}} \boldsymbol{w}_i \cdot \mathbf{F}_{\text{local}, i}^k,
\end{gather}
where $\text{Conv1D}_i$ denotes a 1D convolution with kernel size $i$, $\boldsymbol{w}_i$ are the dynamically generated attention weights.

\noindent \textbf{Inter-Segment Global Relationship Modeling.} To model long-range dependencies, this branch applies self-attention to the segment-aware features $\mathbf{F}_\text{seg}^k$. This relates distant segments, which is vital for assessing properties like overall rhythm, consistency, and coherence throughout the performance:
\begin{equation}
    \small
    \mathbf{F}_\text{inter}^k = \text{Softmax}\left( \frac{\mathbf{W}_\mathcal{Q}^\text{inter}\mathbf{F}_\text{seg}^k \mathbf{W}_\mathcal{K}^\text{inter}(\mathbf{F}_\text{seg}^k)^{\mathrm{T}}}{\sqrt{d}} \right) \mathbf{W}_\mathcal{V}^\text{inter}\mathbf{F}_\text{seg}^k,
\end{equation}
where $\mathbf{W}_\mathcal{Q}^\text{inter}$, $\mathbf{W}_\mathcal{K}^\text{inter}$, and $\mathbf{W}_\mathcal{V}^\text{inter}$ are learnable matrices. The outputs from these three branches are aggregated through element-wise addition and then combined with the initial segment-aware feature $\mathbf{F}_\text{seg}^k$ via a residual connection:
\begin{equation}
    \small
    \mathbf{F}_k = \mathbf{F}_\text{seg}^k + \mathbf{F}_\text{intra}^k + \mathbf{F}_\text{local}^k + \mathbf{F}_\text{inter}^k.
\end{equation}
This multi-granularity modeling enables experts to better understand temporal action structures, capturing fine-grained details and long-range dependencies.

\subsection{Score Prediction and Optimization}
The final mixed feature $\mathbf{F}_{\text{mix}}$, which adaptively integrates knowledge from all experts, is fed into a grade-aware decoder for score prediction. Following recent works \cite{10884538,Xu_2025_CVPR}, this decoder $De(\cdot)$, implemented as a Transformer decoder, uses $M$ learnable queries to generate $M$ distinct grade-aware representations $\{\boldsymbol{r}_m\}_{m=1}^M$. These are processed by an MLP to predict a distribution over $M$ predefined grade levels. The final score $\hat{y}$ is then computed as the expectation over these grades, where the grade values are fixed as $G_m = \frac{m-1}{M-1}$. To encourage expert specialization, we also introduce auxiliary predictions during training: for input action type $k$, the corresponding expert feature $\mathbf{F}_k$ is used to predict an auxiliary score $\hat{y}_k$, incentivizing each expert to learn discriminative patterns.

Our MoAKE framework is optimized end-to-end with a composite loss function, as in \cref{eq:loss}. Specifically, $\mathcal{L}_{\text{score}}$ is a Mean Square Error (MSE) loss that supervises both the final score $\hat{y}$ and auxiliary score $\hat{y}_k$. $\mathcal{L}_{\text{route}}$ is a soft cross-entropy loss that guides the router. Instead of using hard one-hot supervision which forces strict expert isolation, we intentionally employ label smoothing ($\epsilon$) to provide soft supervision during training. \textbf{Crucially, during inference, the router dynamically computes weights based solely on visual features without requiring any action labels.} This encourages the router to discover shared, complementary patterns across different action types, preventing rigid boundaries and fostering positive knowledge transfer. $\mathcal{L}_{\text{div}}$ is a triplet loss on the grade-aware representations $\{\boldsymbol{r}_m\}_{m=1}^M$ to ensure they capture distinct quality levels. The loss components can be rewritten as:
{\allowdisplaybreaks
\begin{gather}
    \small
    \mathcal{L}_{\text{score}} = \text{MSE}(\hat{y}, y) + \text{MSE}(\hat{y}_k, y), \\
    \small
    \mathcal{L}_{\text{route}} = -\sum_{k=1}^{K} \left( c_k(1-\epsilon) + \frac{\epsilon}{K} \right) \log(w_k), \\
    \small
    \mathcal{L}_{\text{div}}=\sum_{m}\left[\max(\text{cos}(\boldsymbol{r}_m, \boldsymbol{r}_j))-\min(\text{cos}(\boldsymbol{r}_m, \boldsymbol{r}_j)) + \alpha\right]_{+},
\end{gather}}
where $y$ is the ground-truth score, $c_k$ is a one-hot vector indicating the ground-truth action type, $\epsilon$ is a label smoothing hyperparameter, $j\ne m$, $\text{cos}\left( \cdot, \cdot  \right)$ is cosine similarity, ${{\left[ \cdot  \right]}_{+}}$ means $\max \left( 0,\cdot  \right)$, and $\alpha$ is a margin.

\section{Experiments}
\textbf{Benchmarks and Metrics.} To rigorously evaluate all-in-one AQA, we establish benchmarks designed to facilitate future research in this direction. These benchmarks are built upon six popular AQA datasets, covering a wide range of action types, semantic complexities, and temporal lengths.
\begin{itemize}
    \item \textbf{All-in-One AQA}: We first adapt three long-term action datasets: Rhythmic Gymnastics (RG) \cite{zeng2020hybrid}, Figure Skating (Fis-V) \cite{16}, and Artistic Swimming (LOGO) \cite{zhang2023logo}. A single model is trained on the combined training sets of these three datasets and evaluated on each test set separately. That is, the testing follows prior works~\cite{Xu_2025_CVPR,CoFInAl}.
    \item \textbf{Zero/Few-Shot AQA}: To evaluate generalization to unseen action types, we use three short-term datasets: Diving (MTL-AQA) \cite{parmar2019and}, Skiing, etc. (AQA-7) \cite{parmar2019action}, and Surgery (JIGSAWS) \cite{jigsaws}. For \textbf{Zero-shot}, the all-in-one model is directly evaluated on target datasets without any parameter updates. For \textbf{Few-shot}, following standard adaptation protocols~\cite{ai2024multimodal}, we perform full-network few-shot fine-tuning using a minimal subset of data. We use 5\% of the training data for MTL-AQA/AQA-7, and 10\% (two samples per class) for the small-scale JIGSAWS dataset. While tuning only the final prototypes might seem intuitive, full fine-tuning is adopted to ensure the shared semantic space properly adapts to the entirely new domain distributions.
\end{itemize}

Following standard AQA evaluation protocols \cite{11024123,yu2021group,ke2024two}, we report Spearman's Rank Correlation (SRCC, $\rho$) and Relative L2-distance (R-$\ell_2$). SRCC measures the monotonic relationship between predicted and ground-truth scores, while R-$\ell_2$ quantifies the numerical prediction error. Higher SRCC and lower R-$\ell_2$ indicate better performance.

\noindent\textbf{Implementation Details.} All experiments were conducted on an RTX 3090 GPU with PyTorch 2.4.1. Following prior works \cite{zhou2023hierarchical,CoFInAl}, we use pre-trained VST \cite{liu2022video} and I3D \cite{i3d} features for the three long-term and three short-term datasets. For RG/Fis-V/LOGO, we extract 68/124/48 snippets of 32 frames. For the short-term datasets, we extract 10 snippets of 16 frames. Feature dimensions $D$ and $d$ are 1024 and 256. We set the number of experts $K\!=\!3$, grade $M\!=\!4$, and segment-aware prototypes $N\!=\!64$. The model is optimized using Adam and a batch size of 32 for 380 epochs. The learning rate is initialized to 3.4e-4 and decayed to 3.4e-6 using a cosine annealing schedule. We apply a dropout rate of 0.3 and a weight decay of 0.0001. The loss weights $\lambda_1, \lambda_2, \lambda_3$ are set to 6, 1, and 1, respectively. The label smoothing parameter $\epsilon$ is 0.3. To avoid overfitting, we adopt a data augmentation strategy where, with 0.6 probability, we replace 10\% of snippets with those from other two types, using 90\% of the original score as a pseudo-label. More details are in the Appendix.

\begin{table*}[!t]
	\centering
	\tabcolsep=2.8pt
    \caption{Comparisons on three long-term benchmarks. \textbf{†} indicates using textual semantics to guide visual modeling, while other methods utilize only the visual modality. The \textbf{bold}/\underline{underline} is the best/second-best results. ``Average'' shows the mean performance across the three actions. One-by-one results are reported from the original papers and prior works \cite{Xu_2025_CVPR,11024123}, with missing metrics reproduced by us.}
	\renewcommand\arraystretch{0.9}
    \resizebox{\linewidth}{!}{
	\begin{tabular}{@{}c|l|cc|cc|cc|cc@{}}
		\toprule
		\multirow{2}{*}{Type}                                                             & \multirow{2}{*}{Methods} & \multicolumn{2}{c|}{Rhythmic Gymnastics {\scriptsize(RG)}} & \multicolumn{2}{c|}{Figure Skating {\scriptsize(Fis-V)}} & \multicolumn{2}{c|}{Artistic Swimming {\scriptsize(LOGO)}} & \multicolumn{2}{c}{\textbf{Average}}     \\ \noalign{\vspace{1pt}} \cline{3-10} \noalign{\vspace{2pt}} 
		&                          & ~~~~~SRCC ↑               & R-${\ell_2}$ ↓              & ~~~SRCC ↑             & R-${\ell_2}$ ↓            & ~~~~~SRCC ↑               & R-${\ell_2}$ ↓              & SRCC ↑       & R-${\ell_2}$ ↓       \\ \midrule
		\multirow{6}{*}{\begin{tabular}[c]{@{}c@{}}One\\      by\\      One\end{tabular}} & GDLT~\cite{xu2022likert} \scriptsize{\color{gray}(CVPR'22)}           & ~~~~0.765                  & \underline{2.401}                 & ~~0.685       & 3.717               & ~~~~0.647                  & 4.148                 & 0.703          & \underline{3.422}          \\
		& HGCN~\cite{zhou2023hierarchical} \scriptsize{\color{gray}(TCSVT'23)}        & ~~~~0.735                  & 4.341                 & ~~0.246                & 12.628              & ~~~~0.671                  & 6.564                 & 0.583          & 7.844          \\
		& T²CR~\cite{ke2024two} \scriptsize{\color{gray}(Inf. Sci.'24)}    & ~~~~0.727                  & 4.054                 & ~~0.652                & 5.113               & ~~~~0.681                  & 5.973                 & 0.688          & 5.047          \\
		& PAMFN~\cite{tip/ZengZ24} \scriptsize{\color{gray}(TIP'24)}         & ~~~~0.711                  & 4.561                 & ~~0.665                & 7.235               & ~~~~0.659                  & 13.979                 & 0.679          & 8.592          \\ 
        & CoFInAl~\cite{CoFInAl} \scriptsize{\color{gray}(IJCAI'24)}     & ~~~~\textbf{0.807}         & 4.028                 & ~~\textbf{0.716}       & \underline{2.875}         & ~~~~0.698                  & \underline{4.019}           & \underline{ 0.744}    & 3.641          \\
        & ASGTN~\cite{10884538} \scriptsize{\color{gray}(TCSVT'25)}       & ~~~~0.799            & 3.027                 & ~~\underline{0.703}          & 3.039               & ~~~~\underline{0.704}            & 5.695                 & 0.739          & 3.920          \\ \noalign{\vspace{1pt}} \cline{1-10} \noalign{\vspace{2.5pt}}
        AinO           & \cellcolor{yellow!10}\textbf{MoAKE} \scriptsize{\color{gray}(This Paper)}             & \cellcolor{yellow!10}~~~~\underline{0.801}                  & \cellcolor{yellow!10}\textbf{1.822}        & \cellcolor{yellow!10}~~0.647                & \cellcolor{yellow!10}\textbf{2.771}      & \cellcolor{yellow!10}~~~~\textbf{0.790}         & \cellcolor{yellow!10}\textbf{2.708}        & \cellcolor{yellow!10}\textbf{0.753} & \cellcolor{yellow!10}\textbf{2.434} \\ \midrule\midrule
		\multirow{9}{*}{\begin{tabular}[c]{@{}c@{}}All\\      in\\      One\end{tabular}} & GDLT~\cite{xu2022likert} \scriptsize{\color{gray}(CVPR'22)}           & ~~~~0.702                  & 2.716           & ~~0.377                & 4.681               & ~~~~0.447                  & 5.499                 & 0.525          & 4.299          \\
        & HGCN~\cite{zhou2023hierarchical} \scriptsize{\color{gray}(TCSVT'23)}        & ~~~~0.507                  & 4.621                 & ~~0.441                & 2.935         & ~~~~0.358                  & 6.181                 & 0.437          & 4.579          \\
        & T²CR~\cite{ke2024two} \scriptsize{\color{gray}(Inf. Sci.'24)}    & ~~~~0.686                  & 3.164                 & ~~0.594                & 3.569               & ~~~~0.704            & 20.828                & 0.664    & 9.187          \\
        & PAMFN~\cite{tip/ZengZ24} \scriptsize{\color{gray}(TIP'24)}         & ~~~~0.683                  & 3.399                 & ~~0.502                & \underline{2.849}               & ~~~~0.690                  & 13.082                & 0.632          & 6.443          \\
        & CoFInAl~\cite{CoFInAl} \scriptsize{\color{gray}(IJCAI'24)}     & ~~~~0.741                  & 2.834                 & ~~0.496                & 3.032               & ~~~~0.603                  & 6.580                 & 0.624          & 4.149          \\
        & ASGTN~\cite{10884538} \scriptsize{\color{gray}(TCSVT'25)}       & ~~~~0.767            & 3.558                 & ~~0.602          & 3.790               & ~~~~0.545                  & 4.809           & 0.649          & 4.052    \\
        & MCMoE~\cite{xu2026mcmoe} \scriptsize{\color{gray}(AAAI'26)}       & ~~~~\underline{0.771}            & 3.426                 & ~~0.625          & 3.608               & ~~~~0.763                  & 4.816           & \underline{0.726}          & 3.950    \\
        & Cond-Base \scriptsize{\color{gray}(Baseline)}    & ~~~~0.748                  & 2.891                 & ~~0.621                & 3.900               & ~~~~\underline{0.769}                  & 4.924                 & 0.718          & 3.905          \\
        & \cellcolor{blue!4}QGVL\textbf{†}~\cite{xu2025quality} \scriptsize{\color{gray}(TMM'25)}    & \cellcolor{blue!4}~~~~0.736                  & \cellcolor{blue!4}\underline{2.240}                 & \cellcolor{blue!4}~~0.592                & \cellcolor{blue!4}4.377               & \cellcolor{blue!4}~~~~0.746                  & \cellcolor{blue!4}\underline{3.393}                 & \cellcolor{blue!4}0.697          & \cellcolor{blue!4}\underline{3.337}          \\
        & \cellcolor{blue!4}MLAVL\textbf{†}~\cite{Xu_2025_CVPR} \scriptsize{\color{gray}(CVPR'25)}    & \cellcolor{blue!4}~~~~0.755                  & \cellcolor{blue!4}2.655                 & \cellcolor{blue!4}~~\underline{0.632}                & \cellcolor{blue!4}2.919               & \cellcolor{blue!4}~~~~0.765                  & \cellcolor{blue!4}5.043                 & \cellcolor{blue!4}0.722          & \cellcolor{blue!4}3.539          \\
        & \cellcolor{yellow!10}\textbf{MoAKE} \scriptsize{\color{gray}(This Paper)}    & \cellcolor{yellow!10}~~~~\textbf{0.801}         & \cellcolor{yellow!10}\textbf{1.822}        & \cellcolor{yellow!10}~~\textbf{0.647}       & \cellcolor{yellow!10}\textbf{2.771}      & \cellcolor{yellow!10}~~~~\textbf{0.790}         & \cellcolor{yellow!10}\textbf{2.708}        & \cellcolor{yellow!10}\textbf{0.753} & \cellcolor{yellow!10}\textbf{2.434} \\ \bottomrule
	\end{tabular}}
	\label{tab:1}
\end{table*}
\subsection{Comparison with State-of-the-Art Methods}
\label{sec:SOTA}
We compare MoAKE with several state-of-the-art (SOTA) AQA works, including visual-only GDLT \cite{xu2022likert}, HGCN \cite{zhou2023hierarchical}, T²CR \cite{ke2024two}, PAMFN \cite{tip/ZengZ24}, CoFInAl \cite{CoFInAl}, and ASGTN \cite{10884538}, and language-guided QGVL~\cite{xu2025quality} and MLAVL~\cite{Xu_2025_CVPR}. Additionally, to evaluate an intuitive all-in-one AQA approach, we construct a strong \textbf{Conditional Baseline} (Cond-Base) featuring a shared backbone with an action classifier and per-action regression heads. Meanwhile, we also compare our model with MCMoE~\cite{xu2026mcmoe}, an AQA model that recently adopted a mixture-of-experts (MoE) architecture. Since this model constructs experts for different modalities, and obtaining consistent modal information across different datasets is challenging, we evaluate MCMoE exclusively in visual scenarios and use visual features at different scales to simulate the multi-source inputs it requires. We evaluate the average performance across four action types on RG, the `Total Element Score (TES)' closely related to visual performance on Fis-V, and the total score on LOGO. For fair comparisons, we re-implement these methods using their official codes under two settings: (1) \textbf{One-by-One}, where a separate model is trained for each dataset, representing the standard paradigm; (2) \textbf{All-in-One}, where a single model is trained on the combined datasets. Notably, as HGCN and ASGTN construct graph nodes based on video length, we incorporate the same segment-aware temporal aggregation to produce fixed-length features, enabling all-in-one training.

\noindent\textbf{Results on All-in-One.} As shown in \cref{tab:1}, MoAKE significantly outperforms all methods in the all-in-one setting. Notably, it surpasses the strong Conditional Baseline (0.753 vs. 0.726 in Avg. SRCC), proving that simply branching at the regression head is insufficient to overcome cross-action semantic gaps; adaptive feature-level routing is essential. Furthermore, MoAKE exceeds SOTA visual-only (MCMoE) and multi-modal (MLAVL) methods by large margins, demonstrating its superior capacity for diverse actions. Remarkably, our all-in-one MoAKE even outperforms SOTA models trained in the standard one-by-one paradigm, with an average SRCC improvement of 1.2\% and a remarkable 28.9\% reduction in R-$\ell_2$. The substantial gain in R-$\ell_2$ highlights that its ability to generate highly accurate scores, not just correct rankings. It is worth noting that both QGVL and MLAVL introduce additional textual semantics to assist visual modeling, which is orthogonal to the main issue of negative transfer across scenes addressed in our work. Consequently, we expect that our method can be further improved by incorporating action knowledge texts as in \cite{Xu_2025_CVPR}.

An interesting observation is that naively applying existing methods to all-in-one training often results in ``negative transfer,'' with performance dropping significantly compared to their one-by-one counterparts (e.g., an average SRCC drop of 10.17\% for baselines on RG). However, in some cases, all-in-one training can bring benefits, such as PAMFN on the LOGO and HGCN on the Fis-V. This suggests that complementary knowledge exists across different action domains, which can be beneficial, especially for datasets with limited samples. This insight motivates our work. In contrast, as shown in \cref{tab:3}, our MoAKE consistently shows positive transfer across all datasets, achieving an average improvement of 4.7\% in SRCC and 34.4\% in R-$\ell_2$. This indicates that our expert-based architecture effectively captures and leverages these complementary cues.

\begin{table}[!t]
  \begin{minipage}[t]{0.515\textwidth}
  \centering
    \caption{Comparisons of generalization performance on three short-term benchmarks. The \textbf{bold} and \underline{underline} indicate the best and second-best results. \textbf{†} indicates using text semantics.}
  \tabcolsep=1pt
  \resizebox{\linewidth}{!}{
  \begin{tabular}{@{}c|l|cc|cc|cc@{}}
  \toprule
  \multirow{2}{*}{Type}                                                     & \multirow{2}{*}{Methods} & \multicolumn{2}{c|}{Diving} & \multicolumn{2}{c|}{Skiing, etc.} & \multicolumn{2}{c}{Surgery} \\ \noalign{\vspace{1pt}} \cline{3-8} \noalign{\vspace{2pt}} 
    &                          & SRCC↑           & R-${\ell_2}$↓          & SRCC↑           & R-${\ell_2}$↓           & SRCC↑         & R-${\ell_2}$↓          \\ \midrule
  \multirow{9}{*}{\begin{tabular}[c]{@{}c@{}}Zero\\      Shot\end{tabular}} & GDLT~\cite{xu2022likert}                     & 0.088                & \underline{4.564}                & -0.024               & 23.615               & \underline{0.133}                & 15.917               \\
  & HGCN~\cite{zhou2023hierarchical}                     & 0.093                & 9.285                & 0.093                & 32.383               & -0.032               & 30.229               \\
  & T²CR~\cite{ke2024two}                     & 0.125                & 6.351                & -0.008               & 19.809               & -0.017               & \underline{15.249}               \\
  & PAMFN~\cite{tip/ZengZ24}                    & 0.090                & 6.768                & 0.013                & 28.522               & 0.001                & 24.232               \\
  & CoFInAl~\cite{CoFInAl}                  & 0.103                & 5.421                & 0.023                & 28.460               & 0.079                & 31.255               \\
  & ASGTN~\cite{10884538}                    & 0.198                & 8.793                & 0.143                & 37.896               & -0.055               & 28.787               \\
  & MCMoE~\cite{xu2026mcmoe}         & 0.201                & 6.174                & 0.136                & 30.479               & 0.101                & 17.987               \\
  & Cond-Base                    & 0.207                & 5.437                & 0.138                & 27.538               & 0.094               & 19.467               \\
  & \cellcolor{blue!4}QGVL\textbf{†}~\cite{xu2025quality}             & \cellcolor{blue!4}0.205                & \cellcolor{blue!4}4.798                & \cellcolor{blue!4}0.133                & \cellcolor{blue!4}22.307               & \cellcolor{blue!4}0.114                & \cellcolor{blue!4}15.387               \\
  & \cellcolor{blue!4}MLAVL\textbf{†}~\cite{Xu_2025_CVPR}              & \cellcolor{blue!4}\underline{0.211}                & \cellcolor{blue!4}5.017                & \cellcolor{blue!4}\underline{0.148}                & \cellcolor{blue!4}\underline{18.249}               & \cellcolor{blue!4}0.122                & \cellcolor{blue!4}16.053               \\
  & \cellcolor{yellow!10}\textbf{MoAKE (Ours)}                     & \cellcolor{yellow!10}\textbf{0.319}       & \cellcolor{yellow!10}\textbf{3.928}       & \cellcolor{yellow!10}\textbf{0.210}       & \cellcolor{yellow!10}\textbf{12.640}       & \cellcolor{yellow!10}\textbf{0.219}       & \cellcolor{yellow!10}\textbf{12.367}      \\ \midrule
  \multirow{9}{*}{\begin{tabular}[c]{@{}c@{}}Few\\      Shot\end{tabular}}  & GDLT~\cite{xu2022likert}                     & 0.605              & 2.319             & 0.463               & 10.426             & 0.602             & 12.568            \\
  & HGCN~\cite{zhou2023hierarchical}                     & 0.387              & 2.894             & 0.383               & 12.913             & 0.498             & 15.097            \\
  & T²CR~\cite{ke2024two}                     & 0.513              & 3.277             & 0.417               & 13.618             & 0.543             & 16.675            \\
  & PAMFN~\cite{tip/ZengZ24}                    & 0.517              & 2.913             & 0.405               & 12.579             & 0.581             & 15.703            \\
  & CoFInAl~\cite{CoFInAl}                  & 0.500              & 3.232             & 0.457               & 13.441             & 0.595             & 16.153            \\
  & ASGTN~\cite{10884538}                    & 0.577              & 2.199             & 0.443               & \underline{10.094}             & 0.576             & \underline{12.082}            \\
  & MCMoE~\cite{xu2026mcmoe}          & 0.598                & 2.311                & 0.452                & 10.732               & 0.603               & 12.447               \\
  & Cond-Base                    & 0.602                & 2.507                & 0.455                & 11.791               & 0.608               & 13.472               \\
  & \cellcolor{blue!4}QGVL\textbf{†}~\cite{xu2025quality}             & \cellcolor{blue!4}0.589                & \cellcolor{blue!4}\underline{2.117}                & \cellcolor{blue!4}\underline{0.469}                & \cellcolor{blue!4}10.132               & \cellcolor{blue!4}0.611                & \cellcolor{blue!4}14.578               \\
  & \cellcolor{blue!4}MLAVL\textbf{†}~\cite{Xu_2025_CVPR}              & \cellcolor{blue!4}\underline{0.618}                & \cellcolor{blue!4}2.244                & \cellcolor{blue!4}0.458                & \cellcolor{blue!4}11.247               & \cellcolor{blue!4}\underline{0.620}                & \cellcolor{blue!4}13.047               \\
  & \cellcolor{yellow!10}\textbf{MoAKE (Ours)}                     & \cellcolor{yellow!10}\textbf{0.664}     & \cellcolor{yellow!10}\textbf{2.058}    & \cellcolor{yellow!10}\textbf{0.513}      & \cellcolor{yellow!10}\textbf{9.138}     & \cellcolor{yellow!10}\textbf{0.651}    & \cellcolor{yellow!10}\textbf{11.168}   \\ \bottomrule
  \end{tabular}}
  \label{tab:2}
  \end{minipage}
  \hfill
  \begin{minipage}[t]{0.475\textwidth}
    \centering
    \caption{Transfer gains ($\Delta$ (\%)) of our MoAKE from one-by-one (ObyO) to all-in-one (AinO). Metrics: SRCC↑ / R-${\ell_2}$↓, `+' denotes improvement.}
    \vspace{-1mm}
    \tabcolsep=3.5pt
    \resizebox{\linewidth}{!}{
    \begin{tabular}{@{}l|cccc@{}}
    \toprule
    Type & RG & Fis-V & LOGO & \textbf{Average}     \\ \midrule
    ObyO        & 0.778 / 4.281             & 0.640 / 2.937             &  0.724 / 3.908         &  0.719 / 3.709        \\
    AinO     & 0.801 / 1.822             & 0.647 / 2.771             & 0.790 / 2.708      & 0.753 / 2.434          \\
    $\Delta$ (\%)          & \textbf{+3.0} / \textbf{+57.4}            & \textbf{+1.1} / \textbf{+5.7}            & \textbf{+9.1} / \textbf{+30.7}        & \textbf{+4.7} / \textbf{+34.4}     \\ \bottomrule
    \end{tabular}}
    \label{tab:3}
  \vspace{2.5mm}
  \centering
  \caption{Ablations of MoAKE. The metrics are reported on the average of three datasets in the all-in-one setting. The {\color{red}red} / {\color{blue}blue} is performance increase / decrease.}
  \vspace{-1mm}
  \tabcolsep=2pt
  \resizebox{\linewidth}{!}{
  \begin{tabular}{@{}lll@{}}
  \toprule
  Methods                                  & SRCC ↑         & R-${\ell_2}$ ↓       \\ \midrule
  Simple Baseline (Single Branch)               & 0.506          & 5.245          \\
  + SATA (Single Branch)                   & 0.607$^{\footnotesize{\color{red}\uparrow 20\%}}$          & 4.057$^{\footnotesize{\color{red}\downarrow 23\%}}$          \\
  + SATA \& AIISRM (Single Branch)       & 0.655$^{\footnotesize{\color{red}\uparrow 8\%}}$          & 3.439$^{\footnotesize{\color{red}\downarrow 15\%}}$          \\
  + MoAKE w/ Specific Adapter              & 0.742$^{\footnotesize{\color{red}\uparrow 13\%}}$          & 2.628$^{\footnotesize{\color{red}\downarrow 24\%}}$          \\
  + MoAKE w/ Shared Adapter (Full Model) & \textbf{0.753}$^{\footnotesize{\color{red}\uparrow 20\%}}$ & \textbf{2.434}$^{\footnotesize{\color{red}\downarrow 23\%}}$ \\ \midrule
  Full Model w/o AIISRM                    & 0.720$^{\footnotesize{\color{blue}\downarrow 4\%}}$          & 2.905$^{\footnotesize{\color{blue}\uparrow 19\%}}$          \\
  Full Model w/o SATA \& AIISRM          & 0.683$^{\footnotesize{\color{blue}\downarrow 2\%}}$          & 3.109$^{\footnotesize{\color{blue}\uparrow 28\%}}$          \\
  Full Model w/o Shared Adapter            & 0.737$^{\footnotesize{\color{blue}\downarrow 2\%}}$          & 2.648$^{\footnotesize{\color{blue}\uparrow 9\%}}$          \\
  Full Model w/o Soft Router               & 0.529$^{\footnotesize{\color{blue}\downarrow 30\%}}$          & 6.081$^{\footnotesize{\color{blue}\uparrow 150\%}}$          \\
  Full Model w/o $\mathcal{L}_{\text{route}}$                   & 0.730$^{\footnotesize{\color{blue}\downarrow 3\%}}$          & 2.952$^{\footnotesize{\color{blue}\uparrow 21\%}}$          \\
  Full Model w/o $\mathcal{L}_{\text{div}}$                      & 0.718$^{\footnotesize{\color{blue}\downarrow 5\%}}$          & 2.888$^{\footnotesize{\color{blue}\uparrow 19\%}}$          \\
  Full Model w/o $\mathcal{L}_{\text{route}}$ \& $\mathcal{L}_{\text{div}}$         & 0.706$^{\footnotesize{\color{blue}\downarrow 6\%}}$          & 3.131$^{\footnotesize{\color{blue}\uparrow 29\%}}$          \\ \bottomrule
  \end{tabular}}
  \label{tab:4}
  \end{minipage}
\end{table}
\noindent\textbf{Results on Zero/Few-Shot.} \cref{tab:2} presents the generalization performance on unseen short-term action datasets. Zero-shot AQA is extremely challenging as the model must infer quality criteria for new actions without any labeled examples. As expected, most SOTA methods struggle, yielding unsatisfactory results. In contrast, MoAKE demonstrates significantly better generalization, outperforming prior methods by a large margin. This is because MoAKE learns more robust and generalizable representations by modeling complementary patterns from diverse long-term actions. Under the few-shot setting, MoAKE's performance improves dramatically, quickly adapting to the new action types and far exceeding the SOTA baselines. This confirms that our expert-based architecture provides a superior foundational representation for downstream adaptation.

\subsection{Quantitative Analysis}
\noindent\textbf{Ablation Studies.} To validate each component in our MoAKE framework, we conduct ablation studies on the all-in-one benchmark. We start with a \textbf{Simple Baseline} model, which consists of token projection, temporal context enhancement layer, adapter and grade-aware regression head. As shown in \cref{tab:4}, we systematically add or remove our key modules: Segment-Aware Temporal Aggregation (SATA), Adaptive Intra- and Inter-Segment Relationship Modeling (AIISRM), the full Mixture of Action Knowledge Experts (MoAKE) architecture, the soft router, the shared adapter, and the loss components.

The results clearly show that each component contributes positively, with the full MoAKE model achieving the best performance. Notably, upgrading from a single-branch baseline to the MoE architecture (``MoAKE w/ Specific Adapter'') yields substantial gains (13\% in SRCC and 24\% in R-$\ell_2$). This confirms that specialized experts effectively capture discriminative patterns, and adaptively fusing their complementary knowledge is crucial for all-in-one AQA. Removing any component leads to a noticeable performance drop, confirming their necessity and effectiveness. Crucially, removing the soft router causes severe degradation (30\% in SRCC and 150\% in R-$\ell_2$). This indicates that naively fusing expert outputs causes feature space confusion due to conflicting action patterns, a core challenge that our adaptive mixing effectively mitigates. Furthermore, even without the routing loss $\mathcal{L}_{\text{route}}$, the model still achieves 0.730 / 2.952 in two metrics. This remains superior to SOTA methods, proving that our framework does not rely on label-conditioned routing, but inherently adapts to diverse action assessment patterns.

\begin{table}[!t]
	\centering
  \caption{Comparisons of computational costs under the all-in-one setting.}
	\tabcolsep=2pt
	\resizebox{1.0\linewidth}{!}{
		\begin{tabular}{@{}c|c|c|c|c|c|c|c|c|c|c@{}}
			\toprule
			Methods & GDLT & HGCN & T²CR & PAMFN & CoFInAI & ASGTN & QGVL & MLAVL & Cond-Base & \textbf{Ours}\\ \midrule
			Params (M) & 3.20 & 4.16 & 0.64 & 18.06 & 5.24       & 6.92      & 5.58   & 3.82 & 3.40 & 8.01  \\
      FLOPs (G) & 0.268 & 0.532 & 0.534 & 2.562 & 0.509       & 0.870      & 1.261   & 0.778  & 0.460 & 1.440  \\
			Avg. ${\rho}$ (↑) & 0.525 & 0.437 & 0.664 & 0.632 & 0.624 & 0.649       & 0.697      & 0.722  & 0.718  & \textbf{0.753}   \\
      Avg. R-${\ell_2}$ (↓) & 4.299 & 4.579 & 9.187 & 6.443 & 4.149       & 4.052      & 3.337   & 3.539 & 3.905   & \textbf{2.434}   \\
		\bottomrule
	\end{tabular}}
	\label{tab:5}
\end{table}
\noindent\textbf{Computational Costs.} \cref{tab:5} compares the number of parameters, FLOPs, and average performance under the all-in-one setting. Benefiting from our tailored lightweight MoE design ($K=3$ experts operating at the macro-action level), MoAKE achieves the best performance while maintaining highly practical computational costs. Compared to heavy multi-modal models (e.g., PAMFN with 18.06M Params/2.562G FLOPs), our MoAKE requires only 8.01M parameters and 1.44G FLOPs. This demonstrates that our framework avoids the massive parameter inflation typical of traditional MoEs in large language models, making it highly efficient and deployable for real-world all-in-one AQA.

\begin{figure}[!t]
  \centering
  \begin{minipage}[t]{0.485\linewidth}
    \centering
    \includegraphics[width=\linewidth]{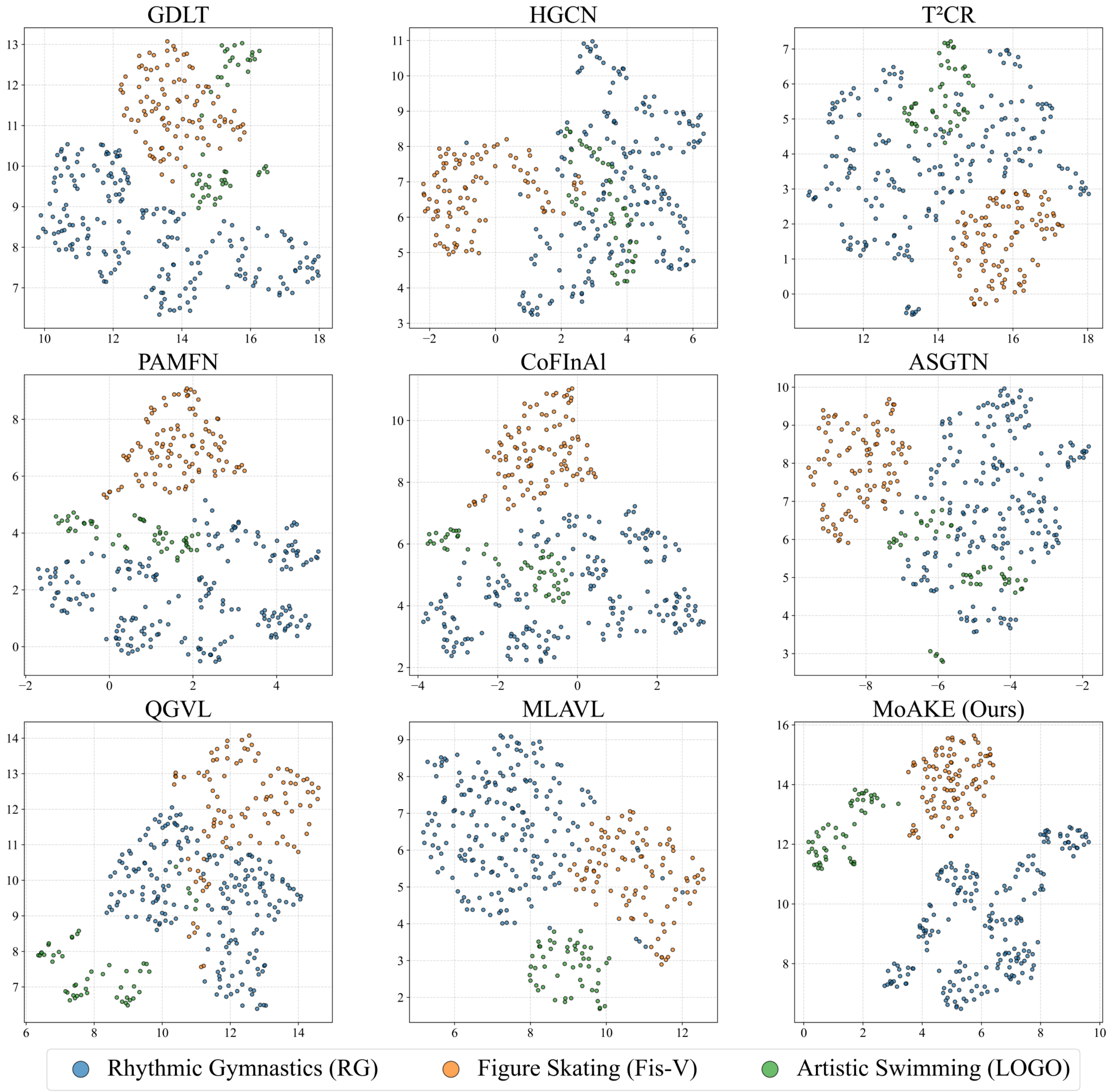}
    \caption{t-SNE visualization of the features learned by different methods for three actions in the all-in-one setting.}
    \label{fig:3}
  \end{minipage}
  \hfill
  \begin{minipage}[t]{0.485\linewidth}
    \centering
    \includegraphics[width=\linewidth]{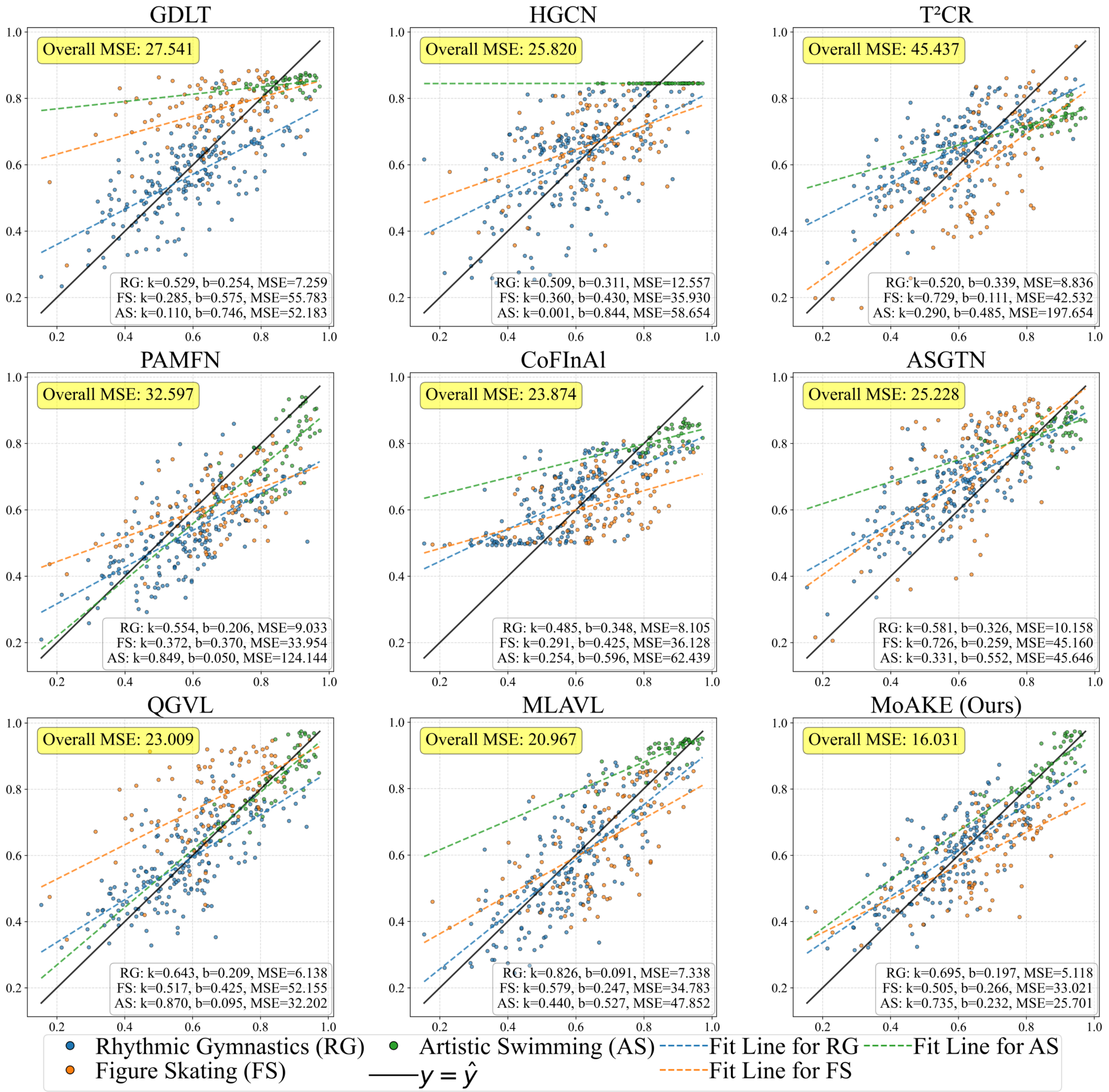}
    \caption{Scatter plots comparing the prediction results of different methods for various actions under all-in-one AQA.}
    \label{fig:4}
  \end{minipage}
\end{figure}
\subsection{Qualitative Analysis}
\noindent\textbf{Semantic Space Visualization.} To qualitatively assess how different methods model the semantic space, we use t-SNE to visualize the video-level features learned by our MoAKE and baselines. As shown in \cref{fig:3}, existing methods fail to distinguish the complex semantics across different types. The features from different action types are poorly separated and intermingled, reflecting their inability to overcome negative knowledge transfer. In stark contrast, MoAKE learns a well-structured feature space where videos of the same action type form distinct, compact clusters, while maintaining clear boundaries and appropriate semantic distances between different types. This intuitively demonstrates that MoAKE effectively learns both shared knowledge and action-specific patterns.

\noindent\textbf{Score Prediction Correlation.} To further analyze prediction results, \cref{fig:4} plots the predicted scores against the ground-truth scores. The scatter plots reveal that MoAKE predicts more accurate scores with tighter correlations to the ground truth compared to other methods, confirming the effectiveness of our unified assessment capability.

\begin{figure}[!t]
  \centering
  \begin{minipage}[t]{0.48\linewidth}
    \centering
    \includegraphics[width=\linewidth]{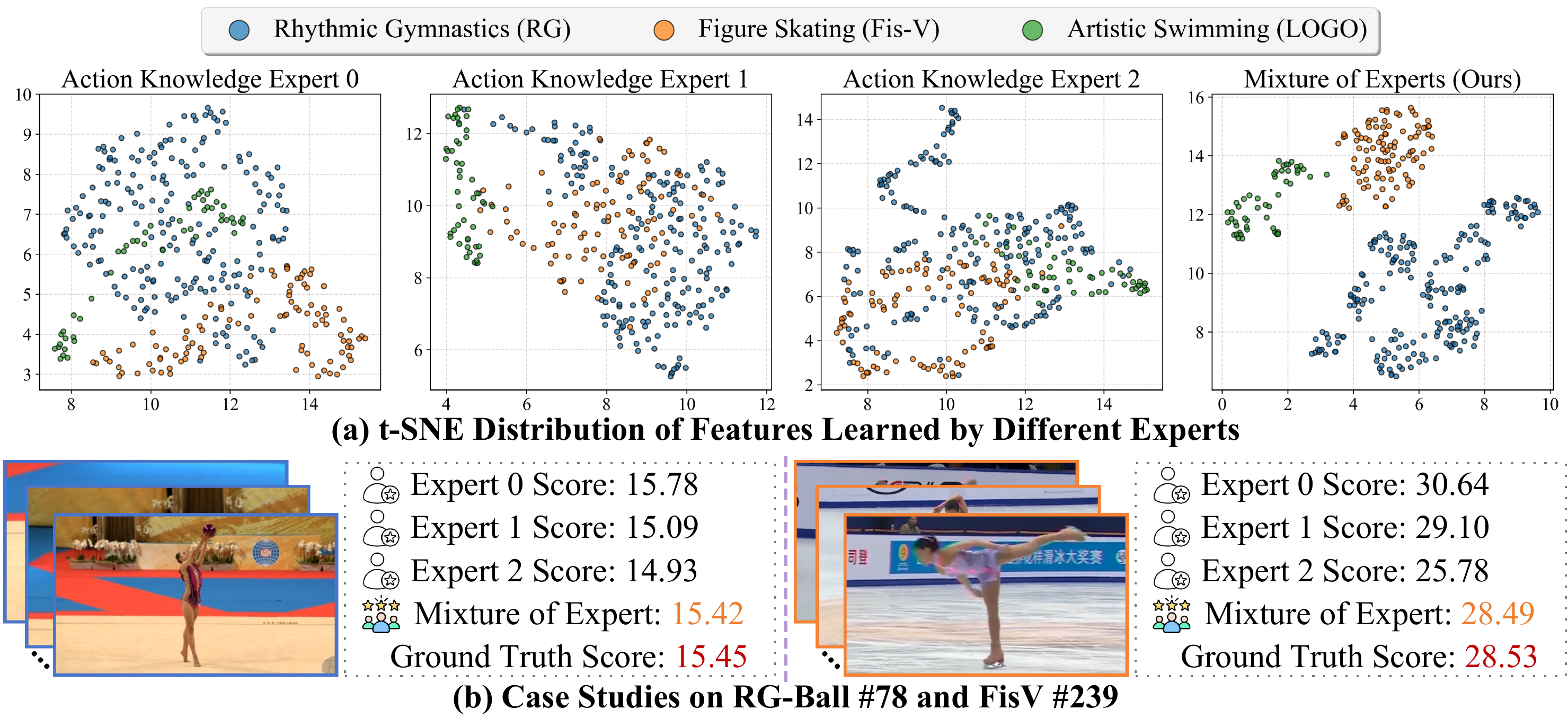}
    \caption{t-SNE distribution of different experts and case studies.}
    \label{fig:5}
  \end{minipage}
  \hfill
  \begin{minipage}[t]{0.51\linewidth}
    \centering
    \includegraphics[width=\linewidth]{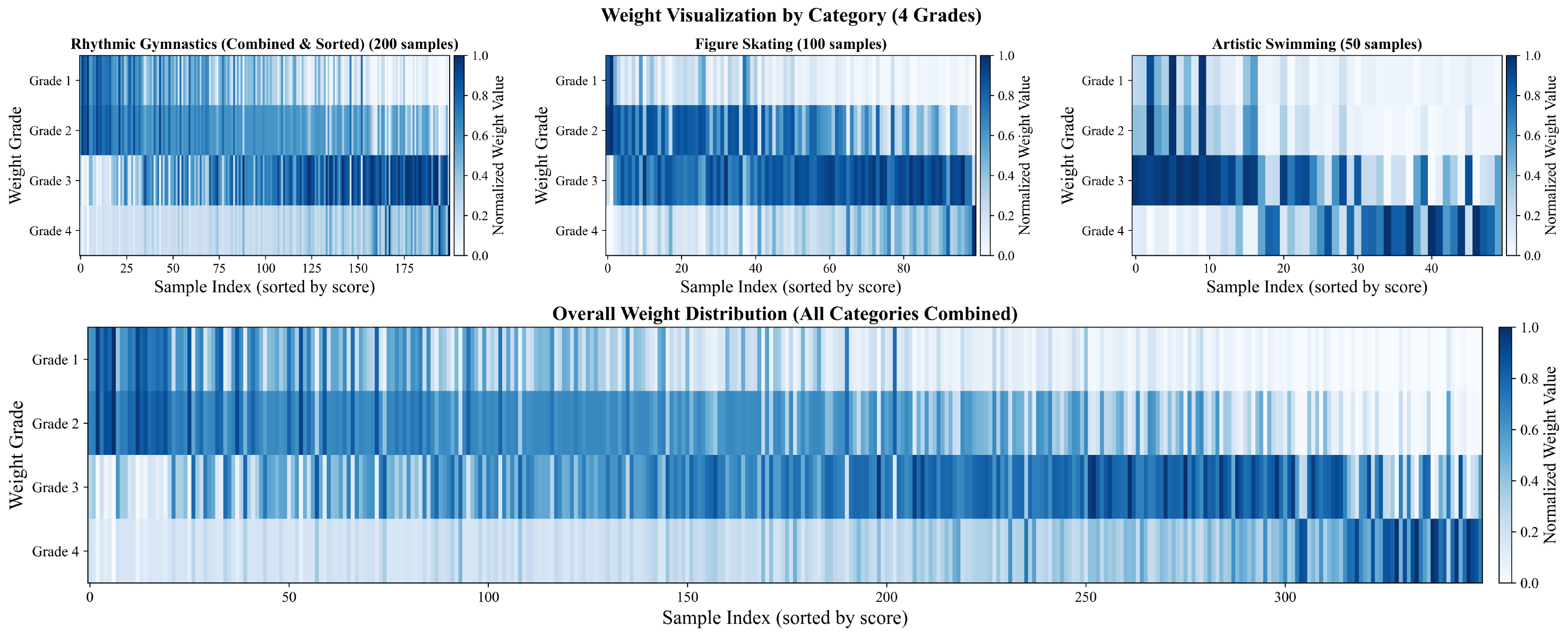}
    \caption{Heatmap of grade pattern weights for each type (top) and all actions (bottom).}
    \label{fig:6}
  \end{minipage}
\end{figure}

\noindent\textbf{Effectiveness of Expert Fusion.} To visualize the mechanism of our soft routing, \cref{fig:5} (a) compares t-SNE distributions of features from individual experts and the final mixed representation. Individual experts specialize in certain action patterns, leading to overlapping clusters. In contrast, the mixed features show clearer inter-class separability. Crucially, as demonstrated by the case studies in \cref{fig:5} (b), while an action-aligned single expert can outperform others on its specific domain, the dynamically mixed representation consistently yields the most accurate predictions. It is important to note that during inference, our soft router relies solely on visual features without accessing any ground-truth action labels. This confirms that our framework does not simply act as a rigid, label-conditioned switch, but rather adaptively extracts and fuses complementary knowledge across all experts to achieve optimal assessment.

\noindent\textbf{Fine-Grained Grade Pattern Weights.} To validate our model's sensitivity to subtle execution details, we visualize the learned grade pattern weights in \cref{fig:6}. The x-axis represents samples sorted by ground-truth scores, and the y-axis denotes progressive quality grades (1-4). A model with strong fine-grained discriminative ability should exhibit a prominent diagonal trend, indicating that higher-scoring samples correctly activate higher-grade patterns. As observed, MoAKE maintains a strict diagonal prominence not only within individual action types (top) but also across the highly complex all-in-one scenario (bottom). This confirms that despite bridging massive cross-action semantic gaps, our framework effectively disentangles nuanced execution differences, demonstrating exceptional fine-grained modeling capabilities for accurate quality assessment.

\section{Conclusion}
This paper introduces a novel framework, Mixture of Action Knowledge Experts (MoAKE), to address the challenging task of all-in-one action quality assessment. We identify the critical limitations of the prevailing ``one-by-one'' paradigm and the ``negative transfer'' issue that arises when naively training a single model on diverse actions. To mitigate this, MoAKE leverages specialized experts to learn discriminative action patterns within a shared semantic space. Complemented by dynamic routing and adaptive relationship modeling, MoAKE transforms the challenge of action diversity into an opportunity for positive knowledge transfer. To validate our approach, we establish the first comprehensive benchmarks for all-in-one and zero/few-shot AQA. Extensive experiments demonstrate that MoAKE not only sets a new state-of-the-art in all-in-one AQA but also exhibits remarkable generalization capabilities. We believe this work paves the way for developing more generic, adaptive, and practical AQA systems.

\noindent\textbf{Future Work.} Although our method shows superior zero-shot results, the current zero-shot scenario remains significantly challenging. Future work could explore ways to further enhance generalization in zero-shot scenarios, such as leveraging self-supervised learning methods or integrating multimodal information to enhance the model's understanding capabilities. Additionally, future work will further expand the all-in-one assessment of more action scenarios.

\section*{Acknowledgements}
This work was supported by the grants from the National Key Research and Development Plan of China (2021YFB3600503), National Natural Science Foundation of China (61972097, U21A20472, 62522102, 62432001), Funds for the Innovation of Policing Science and Technology, Fujian Province (2025YZ040003, 2024YZ040001), Natural Science Foundation of Fujian Province (2025J01536), and the Beijing Natural Science Foundation (L247006).

%
%
\bibliographystyle{splncs04}
\bibliography{main}

\newpage
\appendix

\section{Datasets}
\label{sec:Datasets}
As the first work to systematically explore all-in-one AQA, we establish new benchmarks to facilitate future research. These benchmarks are built upon six popular AQA datasets, covering a wide range of action types, semantic complexities, and temporal lengths.
\begin{itemize}
  \item \textbf{All-in-One AQA}: We first adapt three long-term action datasets: Rhythmic Gymnastics (RG) \cite{zeng2020hybrid}, Figure Skating (Fis-V) \cite{16}, and Artistic Swimming (LOGO) \cite{zhang2023logo}. A single model is trained on the combined training sets of these three datasets and evaluated on each test set separately. That is, the testing follows prior works~\cite{Xu_2025_CVPR,CoFInAl}.
  \item \textbf{Zero/Few-Shot AQA}: To evaluate generalization to unseen action types, we use three short-term datasets: Diving (MTL-AQA) \cite{parmar2019and}, Skiing, etc. (AQA-7) \cite{parmar2019action}, and Surgery (JIGSAWS) \cite{jigsaws}. For \textbf{Zero-shot}, the all-in-one model is directly evaluated on target datasets without any parameter updates. For \textbf{Few-shot}, following standard adaptation protocols~\cite{ai2024multimodal}, we perform full-network few-shot fine-tuning using a minimal subset of data. We use 5\% of the training data for MTL-AQA/AQA-7, and 10\% (two samples per class) for the small-scale JIGSAWS dataset. While tuning only the final prototypes might seem intuitive, full fine-tuning is adopted to ensure the shared semantic space properly adapts to the entirely new domain distributions.
\end{itemize}
\noindent\textbf{Rhythmic Gymnastics (RG).} This dataset encompasses 1,000 rhythmic gymnastics performance videos spanning four distinct apparatus categories: ball, clubs, hoop, and ribbon. Each performance extends approximately 1.6 minutes in duration and is captured at 25 fps. The dataset employs a 4:1 training-evaluation partition, allocating 200 training samples and 50 evaluation samples per apparatus category. We train all apparatus actions uniformly, but during testing, we evaluate different action categories separately and present their average metrics to compare them fairly with existing methods~\cite{xu2022likert,CoFInAl,tip/ZengZ24,10884538}.

\noindent\textbf{Figure Skating Video (Fis-V).} This collection comprises 500 performance videos featuring ladies' singles short programs in competitive figure skating, with each routine lasting approximately 2.9 minutes and recorded at 25 fps. The standard dataset partition designates 400 videos for training purposes and reserves 100 for testing evaluation. Official competition scores are provided for both Total Element Score (TES) and Total Program Component Score (PCS), maintaining consistency with competitive judging standards. Since all-in-one training requires a single score label for each action video, we evaluate the TES score, which is closely related to the athlete's visual action performance, on the Fis-V dataset.

\noindent\textbf{LOGO.} The LOGO collection represents a multi-athlete, extended-duration video dataset containing 150 training instances and 50 testing instances. These recordings capture 26 distinct artistic swimming competitions, each involving 8 performers with an average duration of 204.2 seconds. The dataset incorporates formation annotations to characterize group coordination patterns and provides comprehensive action procedure documentation.

\noindent\textbf{MTL-AQA.} This comprehensive dataset serves as a widely-adopted benchmark in AQA research, featuring 16 distinct diving categories across 1412 total samples. The collection encompasses diverse competition scenarios including male and female participants, individual and synchronized diving events, both 3m springboard and 10m platform competitions. Beyond quality assessment scores, MTL-AQA provides supplementary annotations for action categories and expert commentary.

\noindent\textbf{AQA-7.} This multi-disciplinary sports dataset represents a extensively-utilized resource in AQA research, containing 1189 samples across 7 competitive scenarios: 370 diving samples from 10m platform, 176 gymnastic vault samples, 175 big air skiing samples, 206 big air snowboarding samples, 88 synchronized diving samples from 3m springboard, 91 synchronized diving samples from 10m platform, and 83 trampoline samples.

\noindent\textbf{JIGSAWS.} This surgical skill assessment dataset encompasses three fundamental surgical procedures: ``Suturing (S)'', ``Needle Passing (NP)'' and ``Knot Tying (KT)''. Each video comprises multiple scores under various evaluation criteria, with the aggregate score representing the final assessment. To maintain consistency with prior research~\cite{yu2021group, ke2024two}, we exclusively utilize the left camera perspective and implement a comparable four-fold cross-validation methodology for experimental validation.

\section{More Visualizations}
\label{sec:visuals}
In this section, we provide more visualizations to illustrate the contribution of our proposed method to all-in-one action quality assessment.

\begin{figure*}[!b]
	\centering
	\includegraphics[width=\linewidth]{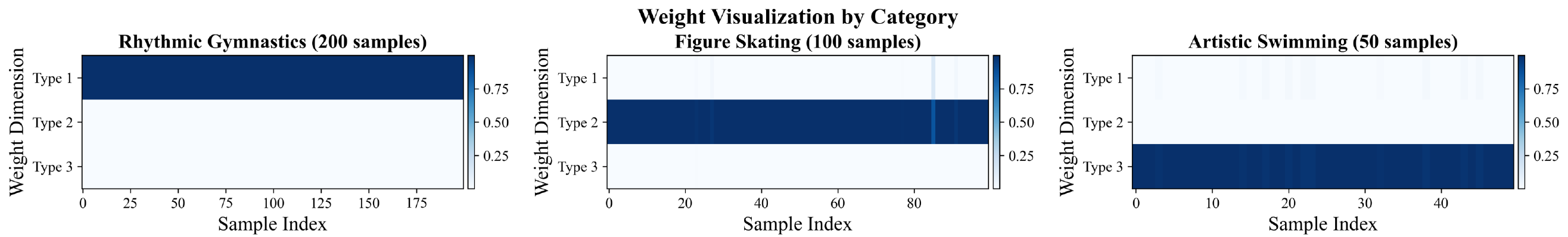}
    \caption{Weight distribution of the soft router $R(\cdot)$. Each subplot corresponds to an action category, where the horizontal axis represents the index of video samples and the vertical axis represents the weight of the corresponding expert.}
	\label{fig:a1}
\end{figure*}
\begin{figure*}[!t]
	\centering
	\includegraphics[width=\linewidth]{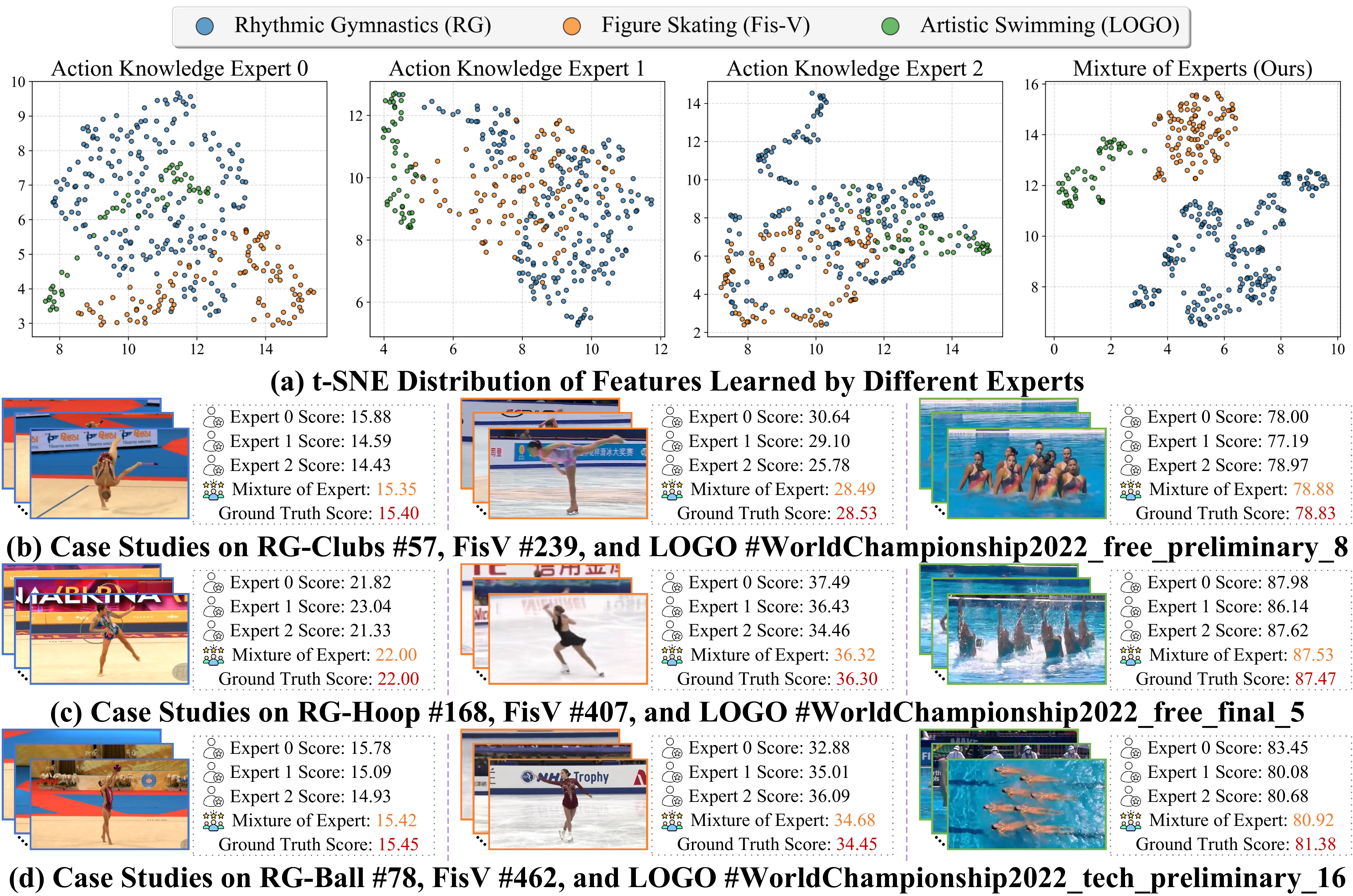}
    \caption{t-SNE distribution of features learned by different experts for three action scenarios and some case studies.}
    \vspace{-8pt}
	\label{fig:a2}
\end{figure*}
\noindent\textbf{Router Weights Distribution.} We show the weight distribution of the soft router $R(\cdot)$ for different input action types on the test datasets in~\cref{fig:a1}. Each subplot corresponds to an action category, where the horizontal axis represents the index of video samples and the vertical axis represents the weight of the corresponding expert. As can be observed, the soft router is capable of dynamically adjusting the weight allocation of experts according to the characteristics of different action categories, accurately selecting the corresponding representative action knowledge experts, thereby achieving a more precise assessment of action quality. This represents the key factor underlying the superior performance of our method in the all-in-one training scenario.

\noindent\textbf{Effectiveness of the Mixture of Experts.} In~\cref{fig:a2}, we expand upon the analysis presented in the main paper by providing more detailed visualizations and case studies. Our MoAKE framework utilizes multiple action knowledge experts to capture distinct action-specific patterns, dynamically fusing their representations via an adaptive routing mechanism. To validate the efficacy of this mixture strategy, we present t-SNE visualizations of the feature distributions learned by individual experts compared to the fused representation from our Mixture-of-Experts (MoE) framework across three action scenarios in~\cref{fig:a2} (a). The first three subplots show features from individual experts, while the rightmost subplot displays the unified representation.

The visualizations reveal that while individual experts specialize in certain action patterns, they exhibit limited discriminative power across diverse action types, leading to overlapping feature clusters. In contrast, our MoE approach yields more discriminative feature embeddings, characterized by enhanced inter-class separability and tighter intra-class clustering. This demonstrates that MoAKE effectively synthesizes complementary knowledge from multiple experts and dynamically adapts to the assessment requirements of various input actions, culminating in superior performance for all-in-one AQA. Furthermore, we provide additional qualitative examples in~\cref{fig:a2} (b-d), which illustrate that different experts focus on distinct aspects of an action. By integrating these diverse perspectives, the MoE model achieves a more comprehensive and accurate final score.

\begin{figure*}[!t]
	\centering
	\includegraphics[width=\linewidth]{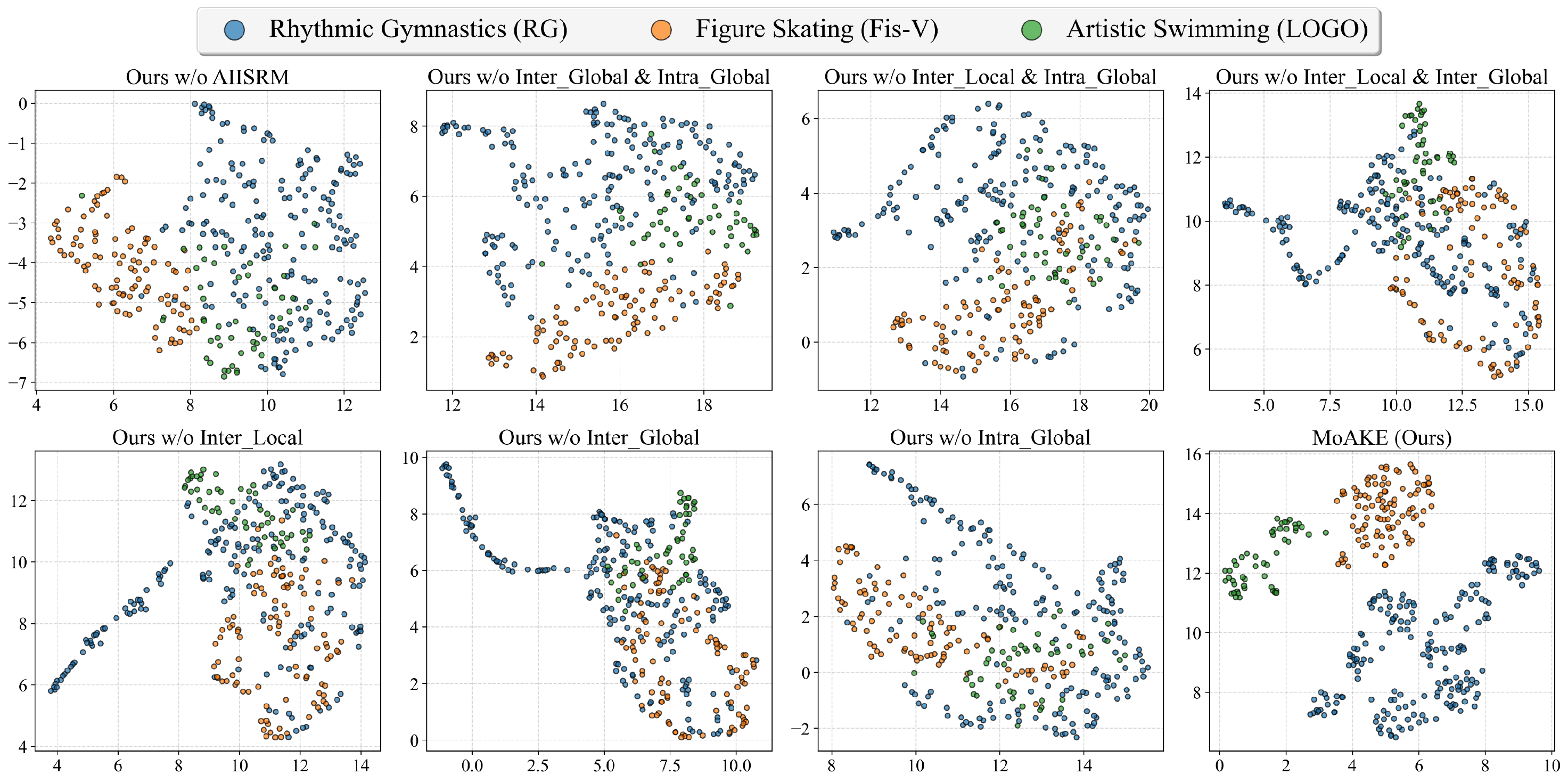}
    \caption{t-SNE visualization of the effects of different relationship modeling branches in our AIISRM module. ``Inter\_Global'', ``Inter\_Local'', and ``Intra\_Global'' correspond to Inter-Segment Global Relationship Modeling, Inter-Segment Local Relationship Modeling, and Intra-Segment Global Relationship Modeling, respectively. The results show that removing any branch leads to chaotic feature distributions, indicating that each branch plays a crucial role in the final feature representation.}
    \vspace{-15pt}
	\label{fig:a3}
\end{figure*}
\noindent\textbf{Effects of the AIISRM.} In our AIISRM module, we employ three distinct relationship modeling branches: Intra-Segment Global Relationship Modeling, Inter-Segment Local Relationship Modeling, and Inter-Segment Global Relationship Modeling. To validate the effectiveness of each branch, we visualize the t-SNE distributions of features learned with and without each branch in~\cref{fig:a3}. The subplots labeled ``Inter\_Global'', ``Inter\_Local'', and ``Intra\_Global'' correspond to the three branches, respectively.

The analysis reveals that removing any single branch leads to a significant degradation in feature distribution, resulting in overlapping clusters and reduced discriminative power. This indicates that each branch plays a crucial role in capturing different aspects of action relationships, and their combination is essential for achieving robust and comprehensive feature representations. The results demonstrate the effectiveness of our AIISRM module in enhancing the quality of action representations through adaptive intra- and inter-segment relationship modeling.

\section{More Ablation Studies}
\label{sec:ablations}
In this section, we further conduct some ablation studies to determine the experimental details. Unless otherwise stated, all ablations are performed on the all-in-one benchmark, containing three long-term action types. We report the average results of three action types.

\begin{table}[t]
    \centering
	\tabcolsep=2pt
    \renewcommand\arraystretch{0.9}
    \caption{Effects of AIISRM Branches. The first row shows the results of our full model, while the other rows show the results of different branch combinations. ``Inter\_Global'', ``Inter\_Local'', ``Intra\_Global'', and ``Intra\_Local'' indicate Inter-Segment Global, Inter-Segment Local, Intra-Segment Global, and Intra-Segment Local Relationship Modeling, respectively.  The {\color{blue}blue} is performance decrease.}
    \resizebox{0.8\linewidth}{!}{
    \begin{tabular}{@{}lll@{}}
    \toprule
    Methods                                  & SRCC ↑         & R-${\ell_2}$ ↓       \\ \midrule
    \textbf{MoAKE (Ours)}              & \textbf{0.753} & \textbf{2.434} \\
    MoAKE w/o AIISRM                  & 0.721$^{\footnotesize{\color{blue}\downarrow 4\%}}$          & 2.907$^{\footnotesize{\color{blue}\uparrow 19\%}}$          \\
    MoAKE w/o ``Inter\_Global'' \& ``Intra\_Global''       & 0.730$^{\footnotesize{\color{blue}\downarrow 3\%}}$          & 2.878$^{\footnotesize{\color{blue}\uparrow 18\%}}$          \\
    MoAKE w/o ``Inter\_Local'' \& ``Intra\_Global''              & 0.736$^{\footnotesize{\color{blue}\downarrow 2\%}}$          & 2.866$^{\footnotesize{\color{blue}\uparrow 18\%}}$          \\
    MoAKE w/o ``Inter\_Local'' \& ``Inter\_Global'' & 0.725$^{\footnotesize{\color{blue}\downarrow 4\%}}$          & 2.827$^{\footnotesize{\color{blue}\uparrow 16\%}}$  \\ 
    MoAKE w/o ``Inter\_Local''                    & 0.744$^{\footnotesize{\color{blue}\downarrow 1\%}}$          & 2.621$^{\footnotesize{\color{blue}\uparrow 8\%}}$          \\
    MoAKE w/o ``Inter\_Global''          & 0.739$^{\footnotesize{\color{blue}\downarrow 2\%}}$          & 2.653$^{\footnotesize{\color{blue}\uparrow 9\%}}$          \\
    MoAKE w/o ``Intra\_Global''            & 0.743$^{\footnotesize{\color{blue}\downarrow 1\%}}$          & 2.749$^{\footnotesize{\color{blue}\uparrow 13\%}}$          \\ \midrule
    MoAKE w/ ``Intra\_Local''               & 0.748$^{\footnotesize{\color{blue}\downarrow 1\%}}$          & 2.548$^{\footnotesize{\color{blue}\uparrow 5\%}}$      \\ \bottomrule
    \end{tabular}}
	\label{tab:a1}
\end{table}
\noindent\textbf{Effects of AIISRM Branches.} Effective AQA requires understanding complex temporal relationships at multiple granularities - both within individual action segments (intra-segment) and across different segments (inter-segment). These relationships are crucial for assessing short-term execution details and long-term action coherence \cite{zhou2023hierarchical,CoFInAl}. To address this need, we propose the novel Adaptive Intra- and Inter-Segment Relationship Modeling (AIISRM) module for each expert. Our AIISRM module incorporates three distinct relationship modeling branches: Intra-Segment Global (``Intra\_Global''), Inter-Segment Local (``Inter\_Local''), and Inter-Segment Global (``Inter\_Global'') Relationship Modeling.

To validate the effectiveness of each branch, we present the results of different branch combinations in~\cref{tab:a1}. We observe that removing any single branch leads to significant performance degradation, showing that each branch plays a crucial role in capturing different aspects of action relationships. Additionally, we also experimented with adding an intra-segment local relationship modeling (``Intra\_Local'') branch, which resulted in a slight performance decrease. This may be attributed to the fact that our Segment-Aware Temporal Aggregation (SATA) already effectively captures local relationships when aggregating segment patterns, making additional local modeling branches potentially redundant. We also provide t-SNE visualization results for different branches in~\cref{fig:a3}, further validating the effectiveness of each branch in capturing action relationships.

\begin{table}[t]
	\centering
	\tabcolsep=5pt
	\renewcommand\arraystretch{0.9}
	\caption{Effects of different loss functions. $\checkmark$ shows that this loss item has been used.}
    \vspace{-5pt}
    \begin{tabular}{@{}cccc|cc@{}}
        \toprule
        $\mathcal{L}_{\text{score}}^{main}$   & $\mathcal{L}_{\text{score}}^{k}$    & $\mathcal{L}_{\text{route}}$     & $\mathcal{L}_{\text{div}}$     & SRCC ↑         & R-${\ell_2}$ ↓     \\ \midrule
		$\checkmark$           &            &          &    & 0.697          & 3.214 \\
		$\checkmark$  & $\checkmark$          &          &    & 0.704          & 3.134 \\
		$\checkmark$    &      & $\checkmark$      &   & 0.708          & 2.961 \\
		$\checkmark$   &     &          & $\checkmark$         & 0.722          & 3.226 \\
		$\checkmark$           &  $\checkmark$          & $\checkmark$        &    & 0.716          & 2.898 \\
		$\checkmark$   & $\checkmark$     &      & $\checkmark$      & 0.724          & 2.951 \\ 
        $\checkmark$   &     & $\checkmark$      & $\checkmark$      & 0.742          & 2.538 \\ 
		$\checkmark$  & $\checkmark$       & $\checkmark$   & $\checkmark$     & \textbf{0.753} & \textbf{2.434} \\ 
		\bottomrule
	\end{tabular}
    \vspace{-5pt}
	\label{tab:a2}
\end{table}
\noindent\textbf{Effects of Different Loss Functions.} Our comprehensive loss function comprises several components: the primary AQA task-specific score fitting loss $\mathcal{L}_{\text{score}}^{main}$, the soft router loss $\mathcal{L}_{\text{route}}$, and the grade pattern diversity loss $\mathcal{L}_{\text{div}}$. Additionally, during training, we perform score regression on features from the discriminative expert corresponding to the current action type, employing $\mathcal{L}_{\text{score}}^{k}$ to ensure that each expert learns distinctive action-specific patterns. Note that $\mathcal{L}_{\text{score}}$ in the main paper is composed of $\mathcal{L}_{\text{score}}^{main}$ and $\mathcal{L}_{\text{score}}^{k}$. To validate the effectiveness of our proposed loss functions, we conduct comprehensive ablation studies on different loss combinations in~\cref{tab:a2}.

The final row shows the results of our full model, which incorporates all loss functions. We observe that each loss component contributes to performance improvement, with the combination of all losses achieving optimal results. Specifically, the score loss $\mathcal{L}_{\text{score}}$ ensures accurate score prediction, the soft router loss $\mathcal{L}_{\text{route}}$ encourages effective expert selection, and the grade pattern diversity loss $\mathcal{L}_{\text{div}}$ promotes diverse grade representations. The auxiliary score regression loss $\mathcal{L}_{\text{score}}^{k}$ further enhances the discriminative power of individual experts. The results demonstrate that our comprehensive loss function effectively guides the model to learn robust and discriminative action representations in shared all-in-one quality spaces.

\noindent\textbf{Effects of Different Number of Grade Patterns $M$.} Following prior works~\cite{xu2022likert,CoFInAl,Xu_2025_CVPR,10884538,11024123}, we adopt state-of-the-art grade-based regression networks that utilize several learnable grade prototypes to aggregate visual features corresponding to different action quality levels. To validate the impact of different numbers of grades on quality assessment, we present results for different numbers of grade patterns $M$ in~\cref{tab:a3}. We observe that performance gradually improves as the number of grade patterns increases, indicating that more grade patterns help capture finer-grained quality distinctions. However, when $M$ becomes excessively large, performance slightly decreases, which may be attributed to overfitting caused by too many grade patterns or the difficulty for neighboring grade patterns to effectively distinguish similar quality information. Considering both performance and computational complexity, we select $M=4$ as the number of grade patterns.

\noindent\textbf{Effects of Different Number of Action Knowledge Experts $K$.} In our MoAKE framework, we employ multiple action knowledge experts to capture diverse action-specific patterns and dynamically fuse their representations through an adaptive routing mechanism. To validate the impact of different numbers of experts on all-in-one AQA, we conduct experiments with varying numbers of experts and present the results in~\cref{tab:a5}. It is noteworthy that when $K=3$, the number of experts aligns with the number of action scenarios in our benchmark, enabling each expert to potentially specialize in action-specific patterns for a corresponding scenario. When $K \neq 3$ and $K > 1$, we employ the diversity loss $\mathcal{L}_{\text{div}}$ to ensure that each expert focuses on distinct action patterns. We observe that as the number of experts increases, performance initially improves, indicating that more experts help capture richer action-specific knowledge. However, when the number of experts becomes excessively large, the performance improvement becomes marginal, which may be attributed to redundancy among experts. Additionally, a larger number of experts leads to significantly increased computational costs. Considering both performance and computational efficiency, we select $K=3$ as the optimal number of action knowledge experts.

\noindent\textbf{Effects of Different Number of Segment-Aware Prototypes $N$.} In our action knowledge expert, we introduce a set of learnable segment queries to capture diverse segment-level action patterns. The number of these segment queries, denoted as $N$, directly influences the granularity of segment representation. To investigate the impact of different values of $N$ on all-in-one AQA, we conduct experiments with varying numbers of segment queries and present the results in~\cref{tab:a4}. We observe that as $N$ increases, performance initially improves, indicating that more segment queries help capture finer-grained action details. However, when $N$ becomes excessively large, performance slightly decreases, which may be attributed to overfitting or increased model complexity. Considering both performance and computational efficiency, we select $N=64$ as the optimal number of segment-aware prototypes. 

\begin{table}[!t]
  \begin{minipage}[t]{0.49\textwidth}
    \centering
    \tabcolsep=2pt
    \caption{Effects of different number of grade patterns $M$.}
    \resizebox{\linewidth}{!}{
    \begin{tabular}{@{}l|ccccc@{}}
        \toprule
        $M$           & 2     & 3     & \textbf{4}              & 5     & 6     \\ \midrule
        SRCC ↑        & 0.735 & 0.744 & \textbf{0.753} & 0.751 & 0.749 \\
        R-${\ell_2}$ ↓ & 2.621 & 2.489 & \textbf{2.434} & 2.456 & 2.478 \\ \bottomrule
    \end{tabular}}
    \label{tab:a3}
  \end{minipage}
  \hfill
  \begin{minipage}[t]{0.49\textwidth}
    \centering
    \tabcolsep=2pt
    \caption{Different number of action knowledge experts $K$.}
    \resizebox{\linewidth}{!}{
    \begin{tabular}{@{}l|ccccc@{}}
        \toprule
        $K$           & 1     & 2     & \textbf{3}              & 4     & 5     \\ \midrule
        SRCC ↑        & 0.655 & 0.738 & \textbf{0.753} & 0.758 & 0.760 \\
        R-${\ell_2}$ ↓ & 3.421 & 2.712 & \textbf{2.434} & 2.412 & 2.387 \\ \bottomrule
    \end{tabular}}
    \label{tab:a5}
  \end{minipage}
\end{table}
\begin{table}[t]
    \centering
    \tabcolsep=2pt
    \caption{Different number of segment-aware prototypes $N$.}
    \resizebox{0.8\linewidth}{!}{
    \begin{tabular}{@{}l|cccccccc@{}}
        \toprule
        $N$           & 16    & 32    & 48    & \textbf{64}             & 80    & 96    & 112   & 128   \\ \midrule
        SRCC ↑        & 0.735 & 0.740 & 0.747 & \textbf{0.753} & 0.750 & 0.747 & 0.744 & 0.745 \\
        R-${\ell_2}$ ↓ & 2.607 & 2.489 & 2.388 & \textbf{2.434} & 2.462 & 2.502 & 2.522 & 2.547 \\ \bottomrule
    \end{tabular}}
    \vspace{-5pt}
    \label{tab:a4}
\end{table}

\begin{table}[!t]
  \begin{minipage}[t]{0.51\textwidth}
    \centering
    \tabcolsep=1pt
    \caption{Comparison of different strategies for handling variable-length videos. Our proposed SATA outperforms conventional methods across different base models.}
    \resizebox{\linewidth}{!}{
    \begin{tabular}{@{}l|cc|cc|cc@{}}
        \toprule
        \multirow{2}{*}{Strategy} & \multicolumn{2}{c|}{MoAKE (Ours)} & \multicolumn{2}{c|}{HGCN~\cite{zhou2023hierarchical}} & \multicolumn{2}{c}{ASGTN~\cite{10884538}} \\ \cmidrule(l{0pt}r{0pt}){2-7}
        & SRCC ↑ & R-${\ell_2}$ ↓ & SRCC ↑ & R-${\ell_2}$ ↓ & SRCC ↑ & R-${\ell_2}$ ↓ \\ \midrule
        Padding\&Truncation & 0.725 & 2.985 & 0.408 & 4.851 & 0.618 & 4.345 \\
        Pooling & 0.731 & 2.854 & 0.423 & 4.789 & 0.630 & 4.276 \\ \midrule
        \textbf{SATA (Ours)} & \textbf{0.753} & \textbf{2.434} & \textbf{0.437} & \textbf{4.579} & \textbf{0.649} & \textbf{4.052} \\ \bottomrule
    \end{tabular}}
    \label{tab:a7}
  \end{minipage}
  \hfill
  \begin{minipage}[t]{0.48\textwidth}
    \centering
    \tabcolsep=6pt
    \renewcommand\arraystretch{0.9}
    \caption{Ablation study on the randomization probability and snippet replacement ratio for our pseudo-label data augmentation strategy.}
    \vspace{-5pt}
    \resizebox{1.0\linewidth}{!}{
    \begin{tabular}{@{}l|ccccccc@{}}
        \toprule
        Probability & 0.1 & 0.2 & 0.4 & \textbf{0.6} & 0.8 & 0.9 \\ \midrule
        SRCC ↑ & 0.745 & 0.748 & 0.750 & \textbf{0.753} & 0.745 & 0.731 \\
        R-${\ell_2}$ ↓ & 2.598 & 2.523 & 2.469 & \textbf{2.434} & 2.502 & 2.634 \\ \midrule \midrule
        Ratio & 5\% & \textbf{10\%} & 15\% & 20\% & 25\% & 30\% \\ \midrule
        SRCC ↑ & 0.749 & \textbf{0.753} & 0.746 & 0.739 & 0.732 & 0.724 \\
        R-${\ell_2}$ ↓ & 2.451 & \textbf{2.434} & 2.498 & 2.571 & 2.645 & 2.719 \\ \bottomrule
    \end{tabular}}
    \vspace{-5pt}
    \label{tab:a9}
  \end{minipage}
\end{table}
\begin{table}[t]
  \centering
  \tabcolsep=7pt
  \caption{Effectiveness of our proposed pseudo-label data augmentation strategy. ``w/o Aug'' and ``w/ Aug'' indicate that the method was trained without and with this strategy, respectively.}
  \resizebox{0.8\linewidth}{!}{
  \begin{tabular}{@{}l|cc|cc@{}}
      \toprule
      \multirow{2}{*}{Method} & \multicolumn{2}{c|}{w/o Aug} & \multicolumn{2}{c}{w/ Aug} \\ \cmidrule(l{0pt}r{0pt}){2-5}
      & SRCC ↑ & R-${\ell_2}$ ↓ & SRCC ↑ & R-${\ell_2}$ ↓ \\ \midrule
      GDLT~\cite{xu2022likert} & 0.512 & 4.481 & 0.525 & 4.299 \\
      HGCN~\cite{zhou2023hierarchical} & 0.423 & 4.789 & 0.437 & 4.579 \\
      T²CR~\cite{ke2024two} & 0.651 & 9.425 & 0.664 & 9.187 \\
      PAMFN~\cite{tip/ZengZ24} & 0.619 & 6.678 & 0.632 & 6.443 \\
      CoFInAl~\cite{CoFInAl} & 0.611 & 4.352 & 0.624 & 4.149 \\
      ASGTN~\cite{10884538} & 0.630 & 4.276 & 0.649 & 4.052 \\
      QGVL~\cite{xu2025quality} & 0.682 & 3.514 & 0.697 & 3.337 \\
      MLAVL~\cite{Xu_2025_CVPR} & 0.705 & 3.761 & 0.722 & 3.539 \\ \midrule
      \textbf{MoAKE (Ours)} & 0.745 & 2.647 & \textbf{0.753} & \textbf{2.434} \\ \bottomrule
  \end{tabular}}
  \vspace{-5pt}
  \label{tab:a8}
\end{table}
\noindent\textbf{Effects of Different Fixed-Length Strategies.} Handling variable-length video inputs is a common challenge in all-in-one action quality assessment. In this work, we introduce Segment-Aware Temporal Aggregation (SATA) to effectively process videos of varying durations. To validate its effectiveness, we compare SATA against several conventional fixed-length strategies, including padding\&truncation and pooling-based sampling. Furthermore, for prior works like HGCN~\cite{zhou2023hierarchical} and ASGTN~\cite{10884538}, which require constructing specific graph structures based on the number of input segments, we also integrated SATA to adapt them to the all-in-one AQA architecture. As shown in~\cref{tab:a7}, the results demonstrate that our proposed SATA significantly outperforms these traditional alternatives across different base models. This indicates that SATA can more effectively capture the rich contextual information from videos of varying lengths, which is crucial for robust all-in-one AQA.

\noindent\textbf{Effectiveness of Pseudo-Label Data Augmentation.} To mitigate overfitting, we employ a data augmentation strategy where, with a probability of 0.6, we replace 10\% of the snippets in a video with snippets from the other two action types, using 90\% of the original score as a pseudo-label. As shown in~\cref{tab:a8}, this strategy is applied to all methods and consistently helps prevent overfitting while further boosting performance. This improvement is likely due to the random mixing of pseudo-labeled data, which enhances the model's adaptability to unseen samples and prevents it from converging to a local optimum for a specific action scenario. Furthermore, we present an ablation study in~\cref{tab:a9} to analyze the impact of the randomization probability and the snippet replacement ratio. Performance steadily improves as the probability increases from 0.1 to 0.6. However, a probability greater than 0.6 or an excessive replacement ratio leads to a significant performance decline, likely because an overabundance of pseudo-data disrupts the model's ability to adapt to the true data distribution.

\section{More Experiment Settings}
\label{sec:settings}
\noindent \textbf{Evaluation Metrics.} Following prior works \cite{yu2021group,11024123,Xu_2025_CVPR}, we adopt the widely used metrics in AQA, including Spearman's Rank Correlation Coefficient (SRCC, ${\rho}$) and Relative L2 distance (R-${\ell_2}$). SRCC measures the rank correlation between the ground-truth score series $\{y\}$ and the predicted score series $\{\hat{y}\}$, while R-${\ell_2}$ measures the relative L2 distance between the ground-truth score ${y}$ and the predicted score ${\hat y}$, which are defined as:
\begin{equation}
    \rho  = \frac{{\sum\nolimits_i {\left( {{y_i} - \bar y} \right)\left( {{{\hat y}_i} - \bar{\hat{y}}} \right)} }}{{\sqrt {\sum\nolimits_i {{{\left( {{y_i} - \bar{y}} \right)}^2}\sum\nolimits_i {{{\left( {{{\hat y}_i} - \bar{\hat{y}}} \right)}^2}} } } }},
\end{equation}
\begin{equation}
    \text{R-}{\ell _2} = \frac{1}{N}{\sum\limits_n^N {\left( {\frac{{\left| {{y_n} - {{\hat y}_n}} \right|}}{{{y_{\max }} - {y_{\min }}}}} \right)} ^2} \times 100.
\end{equation}

\noindent \textbf{Computational Environment.} All experimental evaluations are conducted using a single NVIDIA GeForce RTX 3090 GPU equipped with PyTorch framework version 2.4.1 and an Intel processor operating at 2.40GHz. The CUDA toolkit version employed is 12.4. For reference, training our all-in-one benchmark requires approximately two hours when configured with a batch size of 32 samples and 380 training epochs, utilizing visual representations extracted from pre-trained backbones.

\noindent\textbf{Label Normalization.} Due to varying scoring criteria across different action domains, the score label ranges differ significantly among datasets. Directly utilizing these inconsistent score labels poses substantial optimization challenges for the regression head. To ensure consistent label distributions across different datasets, we apply label normalization to the ground-truth scores, which is consistent with prior works~\cite{xu2022likert,CoFInAl,Xu_2025_CVPR,10884538,11024123}. Specifically, we normalize the scores of each dataset to a range of [0, 1] by dividing by dataset-specific maximum values. In the all-in-one setting, the normalization factors for RG, Fis-V (TES), and LOGO datasets are 25, 45, and 100. In the zero/few-shot setting, MTL-AQA employs a factor of 104.5, while AQA-7's six actions (`diving', `gym\_vault', `ski\_big\_air', `snowboard\_big\_air', `sync\_diving\_3m', and `sync\_diving\_10m') use factors of 110, 25, 50, 50, 110, and 110. For JIGSAWS, the three action categories ('Knot\_Tying', 'Needle\_Passing', and 'Suturing') utilize normalization factors of 25, 25, and 30, respectively. This normalization process is crucial for enabling effective all-in-one training, as it allows the model to learn from diverse datasets with varying score distributions while maintaining numerical stability during optimization.

It is important to note that our entire experimental pipeline, from input processing to model training and output prediction, operates on scores normalized to the [0, 1] range. This normalization is a crucial step that enables unified, all-in-one training across diverse action scenarios and datasets. Consequently, the MSE loss is computed on these normalized scores, preventing datasets with larger original score ranges from dominating the training process. This experimental setup is consistent with all compared prior works, ensuring a fair and direct comparison of performance metrics. For evaluations requiring predictions on the original score scale, such as calculating the MSE difference for each action type as shown in Figure 4 of the main paper, the normalized output scores are de-normalized by multiplying them with their corresponding scaling factors.

\section{Limitations and Broader Impacts}
\label{sec:impacts}
\noindent\textbf{Limitations.} While our MoAKE framework demonstrates strong performance in all-in-one AQA, several limitations warrant discussion. First, although our adaptive routing mechanism effectively selects relevant experts, it may encounter challenges when processing highly ambiguous or novel action types that do not closely align with any existing expert specialization. Expanding the diversity and number of experts could potentially address this limitation, albeit at the cost of increased computational overhead. Future work could explore more sophisticated routing strategies or meta-learning approaches to better handle such edge cases. Second, our current benchmark predominantly focuses on sports and medical actions that have been widely studied in recent literature, which may not fully represent the diversity of action types across other domains. Future research could consider extending to broader action categories and application domains to validate the generalizability and adaptability of our approach.

\noindent\textbf{Broader Impacts.} Our MoAKE framework has the potential to significantly advance the field of action quality assessment by enabling a unified model capable of handling diverse action types. This has important implications for various applications, including sports performance analysis, surgical skill evaluation, and rehabilitation monitoring. By providing accurate and consistent quality assessments across different domains, our approach can facilitate more effective training, feedback, and improvement strategies. However, it is essential to consider ethical implications, such as ensuring fairness and avoiding biases in assessments across different populations or skill levels. Currently, no AQA model is completely accurate and free from bias. Therefore, in practical applications, AQA model results should serve as a reference rather than the sole evaluation criterion. We recommend combining expert opinions and other assessment methods in real-world applications to ensure comprehensive and fair quality evaluation.

\end{document}